\documentclass[utf8]{article}
\usepackage{arxiv}
\usepackage[utf8]{inputenc} 
\usepackage[T1]{fontenc}    
\usepackage{hyperref}       
\usepackage{url}            
\usepackage{booktabs}       
\usepackage{amsfonts}       
\usepackage{nicefrac}       
\usepackage{microtype}      
\usepackage{lipsum}		
\usepackage{graphicx}
\usepackage{natbib}
\usepackage{doi}

\usepackage[english]{babel}
\usepackage{url,hyperref,microtype,subcaption}
\usepackage[onehalfspacing]{setspace}
\usepackage{booktabs} 
\usepackage{longtable} 
\usepackage{amsmath} 
\usepackage{graphicx}
\usepackage{subcaption}
\usepackage{tikz}
\usetikzlibrary{shapes.geometric, arrows.meta, positioning, fit, calc}

\author{
	Islam Akef Ebeid \\
	Division of Computer Science \\
	Texas Woman's University \\
	\texttt{iebeid@twu.edu} \\
	\And
	Haoteng Tang \\
	Department of Computer Science \\
	The University of Texas Rio Grande Valley \\
	\And
	Pengfei Gu \\
	Department of Computer Science \\
	The University of Texas Rio Grande Valley \\
}
\date{}
\begin{document}

	\title{Inferred global dense residue transition graphs from primary structure sequences enable protein interaction prediction via directed graph convolutional neural networks} 

	\maketitle
	\begin{abstract}
		Introduction
		
		Accurate prediction of protein-protein interactions (PPIs) is crucial for understanding cellular functions and advancing the development of drugs. While existing \textit{in-silico} methods leverage direct sequence embeddings from Protein Language Models (PLMs) or apply Graph Neural Networks (GNNs) to 3D protein structures, the main focus of this study is to investigate less computationally intensive alternatives. This work introduces a novel framework for the downstream task of PPI prediction via link prediction.
		
		Methods
		
		We introduce a two-stage graph representation learning framework, $ProtGram-DirectGCN$. First, we developed $ProtGram$, a novel approach that models a protein's primary structure as a hierarchy of globally inferred n-gram graphs. In these graphs, residue transition probabilities, aggregated from a large sequence corpus, define the edge weights of a directed graph of paired residues. Second, we propose a custom directed graph convolutional neural network, $DirectGCN$, which features a unique convolutional layer that processes information through separate path-specific (incoming, outgoing, undirected) and shared transformations, combined via a learnable gating mechanism. $DirectGCN$ is applied to the $ProtGram$ graphs to learn residue-level embeddings, which are then pooled via an attention mechanism to generate protein-level embeddings for the prediction task.
		
		Results
		
		The efficacy of the $DirectGCN$ model was first established on standard node classification benchmarks, where its performance is comparable to that of established methods on general datasets, while demonstrating specialization for complex, directed, and dense heterophilic graph structures. When applied to PPI prediction, the full $ProtGram-DirectGCN$ framework achieves robust predictive power despite being trained on limited data.
		
		Discussion
		
		Our results suggest that a globally inferred, directed graph-based representation of sequence transitions offers a potent and computationally distinct alternative to resource-intensive PLMs for the task of PPI prediction. Future work will involve testing $ProtGram-DirectGCN$ on a wider range of bioinformatics tasks.
	\end{abstract}
	
	\section{Introduction}
	Protein-protein interactions (PPIs) form a network of physical contacts and functional associations mediated by molecular bonds. These interactions are the basis for cellular processes and are collectively referred to as the cellular interactome. Therefore, understanding the underlying mechanisms by predicting valid interactions between proteins is the foundation for many \textit{in-vitro} biomedical endeavors, such as understanding disease mechanisms, drug development and repurposing, and the potential development of futuristic biotechnologies \cite{vidal_interactomenetworkshuman_2011} \cite{scott_smallmoleculesbig_2016}. \textit{In-vitro} protein interaction prediction methods, including Yeast-2-Hybrid screening, co-immunoprecipitation followed by mass spectrometry and affinity purification, have been used to infer empirical evidence of protein association. However, these methods are usually prone to a high rate of false positives and false negatives \cite{rao_proteinproteininteractiondetection_2014}.
	
	Computational methods, also known as \textit{in-silico} and data-driven approaches, have been adopted in life sciences research since at least the seventies \cite{wodak_computeranalysisproteinprotein_1978}. \textit{In-silico} methods help alleviate several of the significant challenges of the \textit{in-vitro} methods mentioned above. The initial stages of drug discovery are heavily based on identifying and confirming valid drug targets, often proteins. This activity is typically protracted, resource-intensive, and time-consuming. Subsequent \textit{in-vitro} screening of drug candidates against potential targets cannot be done efficiently until a validated list of candidate proteins is established. Delays in this upstream target identification task lead to delays in the beginning of extensive \textit{in-vitro} studies \cite{scannell_diagnosingdeclinepharmaceutical_2012}. Therefore, \textit{in-silico} methods, mainly relying on the predictive power of complex machine learning models, are not meant to replace \textit{in-vitro} methods. Instead, they are integrated into the workflow to create a potential pool of valid interactions waiting for wet lab filtering and confirmation, eventually and evidently speeding up the process \cite{vidal_interactomenetworkshuman_2011}.
	
	Recent advancements in machine learning, neural networks, and deep learning approaches have enabled the automation of feature extraction. In addition to embeddings of biological entities into a real vector space, where meaningful algebraic operations can be performed on the learned vectors representing individual residues or proteins. The advent of the transformer architecture \cite{vaswani_attentionallyou_2017} has surpassed sequence-to-sequence models, particularly recurrent neural networks (RNNs) like the Long Short-Term Memory (LSTM) model \cite{hochreiter_longshorttermmemory_1997}. These models have been relied on in protein sequence modeling and representation learning \cite{cho_learningphraserepresentations_2014}. The input protein sequence is usually tokenized at different levels or granularity, such as representing single amino acids as words or a group of residues as k-mers \cite{guo_usingsupportvector_2008}. These models, although efficient in processing short-term dependencies, have demonstrated a limited understanding of context incorporation in language modeling. Though, that contextual understanding has had glimpses in non-recurrent language-based neural networks like Word2Vec \cite{mikolov_distributedrepresentationswords_2013}, where the goal becomes incorporating context via a binary negative loss function that classifies in and out of context window words to the current word. The introduction of the attention mechanism and the transformer architecture, combining both sequence-to-sequence modeling and contextual encoding \cite{vaswani_attentionallyou_2017}, has increased the predictive power of language models by orders of magnitude on multiple tasks. Subsequently that has contributed to the proliferation of different designs and architectures like BERT (encoder only) \cite{devlin_bertpretrainingdeep_2019}, T5 (encoder-decoder) \cite{raffel_exploringlimitstransfer_2023}, and GPT (decoder only) \cite{brown_languagemodelsare_2020}. That, however, comes at a significant computational and environmental cost, due to the increased reliance on training data for these models, as well as the near-linear correlation between a model's predictive power and the number of parameters present in the network. 
	
	Transformer-based architectures have had great success in adoption in domain-specific tasks via fine-tuning; for example, BioBERT \cite{lee_biobertpretrainedbiomedical_2020} fine-tunes BERT over the corpus of PubMed metadata and available full-text on multiple tasks. One of the tasks is biological named entity recognition and extraction for names of diseases, genes, proteins, species, and drugs. The protein embeddings extracted from BioBERT can provide encoded contextual meaning in downstream tasks, such as protein interaction prediction or gene identification. In drug development, particularly in protein-protein interaction prediction (PPI), advanced models have been applied at multiple levels. For example, in reinforcement learning, a prominent recent advancement is Google’s AlphaFold \cite{jumper_highlyaccurateprotein_2021}. This complex model aims to predict protein 3D structures from primary sequences, a central challenge in biomedical informatics known as protein folding prediction. Predicted 3D structures are often utilized in frameworks that aim to predict protein interactions from all levels of protein structure representation via combining features from the primary, secondary, and tertiary structures in addition to topological features from protein-protein interaction networks \cite{zhou_graphneuralnetwork_2022} \cite{jha_predictionproteinprotein_2022}. However, the most common approaches in \textit{in-silico} PPI prediction are primary structure sequence-based methods, where the sequential one-dimensional nature of individual amino acids and residue-level representations lend themselves to modern language modeling. The core lies in the context encoded in the transition probabilities between residues due to the relative simplicity of the input data.
	
	Protein sequences can be conceptualized as a sequence of amino acids (or peptides), analogous to sentences being sequences of words. This analogy allows for the application of large language modeling techniques. For instance, ProtBert \cite{elnaggar_prottransunderstandinglanguage_2022} applies the BERT architecture \cite{devlin_bertpretrainingdeep_2019} to primary protein structures, yielding accurate models on downstream tasks that generate protein-level vector representations at the residue and protein levels. Moreover, work like \cite{sledzieski_dscripttranslatesgenome_2021} D-SCRIPT relies on a PLMs to predict spatial protein interaction contact maps. The model was evaluated on per organism protein-based PPI prediction task yielding positive results. Building on these results, the model provides functionally informative predictions and yields more coherent gene clusters. The predicted contact maps significantly overlap with the true 3D structure contacts, despite being trained solely on sequence data. The common aspect of all of these models is that; first the output embeddings is typically pooled to produce per-protein embeddings that capture sequential features enabling higher predictive power in downstream tasks \cite{elnaggar_prottransunderstandinglanguage_2022}. Second, primary structure sequences themselves appear to encode more than the obvious, even with relatively limited data availability.
	
	However, even the most advanced modern approaches have had several problems.
	\begin{itemize}
		\item The limited context window size must be larger to capture longer-range dependencies beyond the immediate neighborhood which significantly increases their need for labeled training data.
		\item Increasing the number of layers, blocks and attention heads can lead to a significant (potentially exponential) increase in the number of parameters, demanding more computational resources.
		\item The models are generally highly sensitive to training data quality, diversity, volume and availability.
	\end{itemize}
	
	Here, we propose a novel approach to modeling protein primary structure sequences that partially overcomes some of these limitations. We cast the sequences as random walks sampled according to transition probabilities within a directed n-gram graph $G_{n}$ of amino acids. The directed graph $G_{n}$ is inferred from a database of curated protein sequences (UniProt) \cite{theuniprotconsortium_uniprotuniversalprotein_2023}. Then, a custom-directed graph convolution neural network, $DirectGCN$, learns the dense relationships of the transitions between the n-grams. The learned representations are then evaluated on a PPI link prediction task and compared with other established models to establish the method's validity. 
	
	Our approach overcomes the need for a context window \cite{elnaggar_prottransunderstandinglanguage_2022} where the computational limitations that contribute to limited context windows are only applied to a limited dense graph of n-grams. The first-order neighborhood of a spectral graph convolution operator approximation \cite{kipf_semisupervisedclassificationgraph_2017} has a limited effect on the output compared to the sizes of a well-capturing context window in a large language model. In addition this approach in modeling the sequences reduces the number of parameters significantly as the directed graph convolution network operates on a limited unique vocab nodes. In addition to the ability to learn complex encoding from limited training data as the n-gram graph with different levels can act as a data augmentation mechanism if full sequence databases are not available. In addition this bottom-up approach ensures that the limitations are only applied to the lowest level of representation where reducing noise at that level reverberates at higher n-gram level, in addition to overcoming the need for intensive computational power.
	
	From a biological standpoint the specific transition sequence of amino acids via their side chains or R groups determines how a polypeptide chain will fold. Hydrogen bonds, ionic bonds, and hydrophobic interactions generally drive the folding. And in the process, the local secondary structure, including alpha and beta helices, eventually creates binding sites essential for forming subunit proteins or interacting with other molecules. Hence, our intuition is that the primary structure sequences and the transition frequencies between residues holds enough signal power that can inform downstream not only the 3D tertiary structure of the protein but also tell the possible interactions with other proteins or the quaternary structure \cite{dill_proteinfoldingproblem50_2012} \cite{perkins_transientproteinproteininteractions_2010} \cite{anfinsen_principlesthatgovern_1973}.
	
	Accordingly we hypothesize that the global directed dense graph of n-grams $G_{n}$ encodes the potential relationships between proteins, and that learning accurate vector representations of $G_{n}$ not only provides promising performance if further developed compared to PLMs in the task of PPI link prediction but also offers a method to generate protein embeddings on the fly without the need to store per-protein embeddings nor to fine tune hefty pretrained models.
	
	Here we are trying to answer the following research questions:
	\begin{itemize}
		\item Does learning representations of proteins from the embedded, inferred directed graph of n-grams $G_{n}$ encode valid associations between proteins? 
		\item Is our $ProtGram-DirectGCN$ model credible and valid?
		\item What is the predictive power of our hierarchical feature based n-gram representation $G_{n}$?
		\item Is the performance of the $ProtGram-DirectGCN$ model comparable to PLMs? And what are the implications of that?
	\end{itemize}
	
	\section{Methods}
	This section details the design, development, and evaluation of our model $ProtGram-DirectGCN$ model, a directed graph convolutional network tailored for learning representations from dense, directed, and weighted graphs. The primary motivation for this model arises from the need to effectively process a global, complex, dense, and heterophilic graph of n-grams, $G_{n}$, constructed from large-scale protein sequence data where capturing directionality and transition frequencies is paramount for deriving meaningful biological insights.
	
	\subsection{$ProtGram$}
	Our approach treats individual proteins, identified via the comprehensive database UniProt \cite{theuniprotconsortium_uniprotuniversalprotein_2023}, as distinct entities. These proteins form the nodes $V$ in a high-level biological interaction graph canonized in the database BioGRID \cite{oughtred_biogriddatabasecomprehensive_2021} as $G_{PPI}=(V,E_{PPI})$, where edges $E_{PPI}$ represent observed interactions. The main objective is to solve the link prediction problem within $G_{PPI}$. This task is inherently difficult because real-world PPI networks are extremely sparse. The probability of a randomly chosen pair of nodes having a link is given by the graph density, $\frac{2m}{n(n-1)}$, where $n$ is the number of proteins and $m$ is the number of interactions. In typical biological networks, this value is very low, meaning the number of non-interacting pairs vastly exceeds the number of known interactions and creates a severe class imbalance.
	
	A foundational aspect of our methodology is the detailed representation of individual protein sequences using a hierarchy of n-gram graphs. For a given n-gram size $k$, each protein $P_i$ is defined by its primary amino acid sequence $R_i = (r_1, r_2, \ldots, r_L)$. We model the sequence probability under a $k$-th order Markov assumption, where the probability of an amino acid depends on the preceding $k-1$ residues:
	$$ P(R_i) \approx \prod_{j=k}^{L} P(r_j | r_{j-k+1}, \dots, r_{j-1}) $$
	The conditional probabilities are estimated from a large corpus based on the frequency of k-gram occurrences:
	$$ P(r_j | r_{j-k+1}, \dots, r_{j-1}) = \frac{C(r_{j-k+1}, \dots, r_{j-1}, r_j)}{C(r_{j-k+1}, \dots, r_{j-1})} $$
	where $C(\cdot)$ denotes the count of a particular subsequence in the corpus.
	
	We use this principle to conceptualize all protein sequences through a global, directed, and dense graph of k-grams, denoted $G_{k}=(V_{k}, E_{k})$. Here, $V_{k}$ is the finite set of unique k-gram types observed in the corpus. $E_{k}$ represents directed transitions between these k-grams, where an edge $(u, v)$ from k-gram $u$ to $v$ exists if $v$ can be formed by shifting a one-residue window over $u$. For example, for $k=3$, an edge exists from 'ACG' to 'CGT'. Each edge $(u, v)$ is assigned a weight $w_{uv}$ corresponding to the observed frequency of this transition across the entire corpus.
	
\begin{figure}
		\centering
		\includegraphics[width=0.5\linewidth]{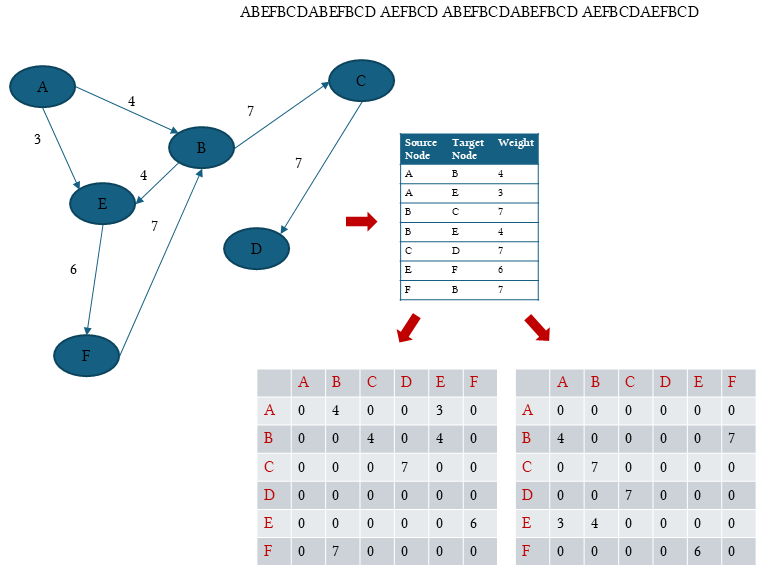}
		\caption{An example of dummy sequences separated by a space indicating multiple proteins. The figure shows how the transitions between the residues or characters are conceptualized as a directed, weighted, dense graph where the weights are the transition frequencies calculated as counts or probabilities. In addition, the figure shows how we split the directed adjacency into an $A_{in}$ and an $A_{out}$. Later, we describe how to overcome the non-hermitian nature of these two matrices to make them suitable for graph convolutional neural networks.}
		\label{fig:1}
	\end{figure}

	A valid protein sequence $R = (r_1, r_2, \ldots, r_L)$ is thus viewed as a specific path or random walk of length $L-1$ on $G_{n}$ on this hierarchy of n-gram graphs. This conceptualization aligns with the idea that protein sequences can be seen as generated from a 'source graph' of amino acid symbols via a probabilistic random walk process. To illustrate this, we can consider the base case of this hierarchy where $n=1$ (a graph of single residues), which corresponds to a first-order Markov process. In this case the probability of observing a particular sequence $R$, given its starting residue $r_1$ and the transition probabilities derived from $G_{n}$, can be formulated. If $P(r_{j+1}|r_j) = \frac{w_{r_j r_{j+1}}}{\sum_k w_{r_j k}}$ is the transition probability from n-gram $r_j$ to $r_{j+1}$ given normalized edge weights, then the probability of the sequence $R$ is $P(R | r_1, G_{n}) = \prod_{j=1}^{L-1} P(r_{j+1}|r_j)$. This probabilistic view, rooted in the empirically derived $G_{n}$, allows for a nuanced understanding of sequence validity, likelihood, and structure. We aim not to view the amino acid sequence representation as a mere Markovian sequence but also to consider the existence of different relationships between a residue and many other residues. The directed nature of $G_{n}$ is crucial, naturally modeling the $N-to-C$ terminus directionality of polypeptide chains and the inherent asymmetry of residue relationships. See figure \ref{fig:1}.
	
	Our custom Directed Graph Convolutional Network $DirectGCN$ is specifically designed to learn from these $G_{n}$ n-gram graphs. Graph Neural Networks (GNNs) are architectures adept at learning node representations by iteratively aggregating information from neighborhoods, also known as message passing \cite{scarselli_graphneuralnetwork_2009}. GNNs can be applied to two different graph domains. The first is spatial and the second is spectral. Spatial models perform message passing across the vertices via direct pass $\rightarrow$ aggregate $\rightarrow$ update computation. While spectral methods rely on operating on the adjacency matrix directly by approximating the convolutional operation. A foundational spectral GNN is the Graph Convolutional Network (GCN) \cite{kipf_semisupervisedclassificationgraph_2017}, whose layer typically updates node features according to $H^{(l+1)} = \sigma\left(\tilde{D}^{-1/2} \tilde{A} \tilde{D}^{-1/2} H^{(l)} W^{(l)}\right)$. Here, $H^{(0)}$ would be initial features for n-gram types in $V_{n}$, $\tilde{A}$ is the adjacency matrix of $G_{n}$ with added self-loops, $\tilde{D}$ is its corresponding diagonal degree matrix for normalization, $W^{(l)}$ is a trainable weight matrix, and $\sigma$ is a non-linear activation function.
	
	Standard GCNs are primarily for undirected graphs assuming symmetric adjacencies. However multiple works have explored applying GCNs to directed graphs for a complete review of directed GCN methods please see the supplementary material. Our $DirectGCN$ adapts this framework to effectively learn from the directed and weighted edges of $G_{n}$, separating information flow based on edge exitance, directionality and the homophily property. Our model learns rich embeddings $V_{n,k} \in \mathbb{R}^{d'}$ for each n-gram type $k$. Rich feature vectors for entire protein sequences $P_i$ are generated by aggregating their constituent n-grams' embeddings learned by $ProtGram-DirectGCN$. These protein-level feature vectors are assembled via an attention mechanism for the downstream PPI link prediction task within $G_{PPI}$. Next we will describe the architecture of the $ProtGram-DirectGCN$ model.
	
	\subsection{$DirectGCN$}
	\subsubsection{Propagation Matrix Formulation}
	The graph structure is initially captured by a raw weighted adjacency matrix $A_{raw} \in \mathbb{R}^{N \times N}$, where $(A_{raw})_{uv} = w_{uv}$. From this, we define the out-degree weighted adjacency matrix $A_{out}^{(w)} = A_{raw}$ and the in-degree weighted adjacency matrix $A_{in}^{(w)} = A_{raw}^T$. In addition we also generate the structural symmetric undirected adjacency $A$. A key component of $DirectGCN$ is a specific preprocessing step for these adjacency matrices, designed to create stable and informative propagation matrices. For a given weighted adjacency matrix $A^{(w)}$ (either $A_{out}^{(w)}$ or $A_{in}^{(w)}$), we first compute its row-normalized counterpart $A^{(n)} = D^{-1}A^{(w)}$, where $D$ is the diagonal out-degree matrix. To overcome the non-hermitian nature of $A^{(n)}$, we compute its symmetric-like ($S$) and skew-symmetric-like ($K$) components:
	\begin{equation}
		S = \frac{A^{(n)} + (A^{(n)})^T}{2} \quad \text{and} \quad K = \frac{A^{(n)} - (A^{(n)})^T}{2} 
	\end{equation}
	The final propagation matrix $\mathcal{A}$ is derived from the element-wise magnitude of these components, with an added identity matrix $I$ for self-loops:
	\begin{equation}
		\mathcal{A} = \sqrt{S^2 + K^2 + \epsilon} + I 
	\end{equation}
	where the square operations are element-wise and $\epsilon$ is a small constant (e.g., $1 \times 10^{-9}$) for numerical stability. This process yields two distinct propagation matrices, $\mathcal{A}_{out}$ and $\mathcal{A}_{in}$, which are used for message aggregation in the convolutional layers. This construction aims to capture both the symmetric and anti-symmetric aspects of the directed relationships, offering a more robust representation of directed influence in addition to the structural path $\mathcal{A}$.
	
	\subsubsection{Propagation Layer}
	Given node features $H^{(l)} \in \mathbb{R}^{N \times F^{(l)}}$ at layer $l$, the layer computes the features for the next layer by processing information through 3 distinct channels: incoming, outgoing, and undirected. Each channel combines a standard graph convolutional message passing operation with a feed forward layer as final feature transformation.
	For the incoming path, the aggregated message is a combination of a propagated component and a shared feature transformation:
	\begin{equation}
		H_{in}^{(l+1)} = \left( \mathcal{A}_{in} (H^{(l)} W_{main,in}^{(l)}) + b_{main,in}^{(l)} \right) + \left( H^{(l)} W_{shared}^{(l)} + b_{shared,in}^{(l)} \right)
	\end{equation}
	Similarly, for the outgoing path:
	\begin{equation}
		H_{out}^{(l+1)} = \left( \mathcal{A}_{out} (H^{(l)} W_{main,out}^{(l)}) + b_{main,out}^{(l)} \right) + \left( H^{(l)} W_{shared}^{(l)} + b_{shared,out}^{(l)} \right)
	\end{equation}
	And for the undirected path, using a standard symmetrically normalized adjacency matrix $\tilde{A}_{undir}$:
	\begin{equation}
		H_{undir}^{(l+1)} = \left( \tilde{A}_{undir} (H^{(l)} W_{main,undir}^{(l)}) + b_{main,undir}^{(l)} \right) + \left( H^{(l)} W_{shared}^{(l)} + b_{shared,undir}^{(l)} \right)
	\end{equation}
	In addition we model the idea of positional encoding which ensures that the model has some notion of time and sequence. We do that by adding a non transformed learnable embeddings layer that gives each node (n-gram) its positional identity:
	\begin{equation}
		B_{const}^{(l)} \in R^{n \times d}
	\end{equation}
	where $W_{main,*}$ are path-specific weight matrices and $W_{shared}$ is a single weight matrix shared across all three paths, acting on the original node features. In addition $d$ is a chosen dimension. These 3 processed signals are then combined using a learnable, node-wise gating mechanism to control the flow of information of each path. alongside a separate learnable feature vector that captures the node positional identity in the graph. Eventually the model resembles an algebraic multivariate first order polynomial linear combination of features that represent separate yet integrated graph properties $aX + bY + cZ + d$:
	\begin{multline}
		H_{\text{pre-activation}}^{(l+1)} = \left( C_{\text{undir}}^{(l)} H_{\text{undir}}^{(l+1)} \right) + \left( C_{\text{in}}^{(l)} H_{\text{in}}^{(l+1)} \right)
		+ \left( C_{\text{out}}^{(l)} H_{\text{out}}^{(l+1)} \right) + B_{\text{const}}^{(l)}
	\end{multline}
	where $C_*^{(l)}$ are the learnable gating vectors that facilitate understanding the importance of the contribution of each path in the learning. Finally, a residual connection is added before applying a Leaky ReLU activation function:
	\begin{equation}
		H^{(l+1)} = \sigma_{LReLU}\left(H_{pre-activation}^{(l+1)} + H^{(l)}W_{res}^{(l)}\right)
	\end{equation}
	where $W_{res}^{(l)}$ is a linear projection for the residual connection if feature dimensions change.
	
	\subsubsection{Model Architecture}
\begin{figure*}[t!]
		\centering
		\begin{tabular}{c}
			\includegraphics[width=0.7\linewidth]{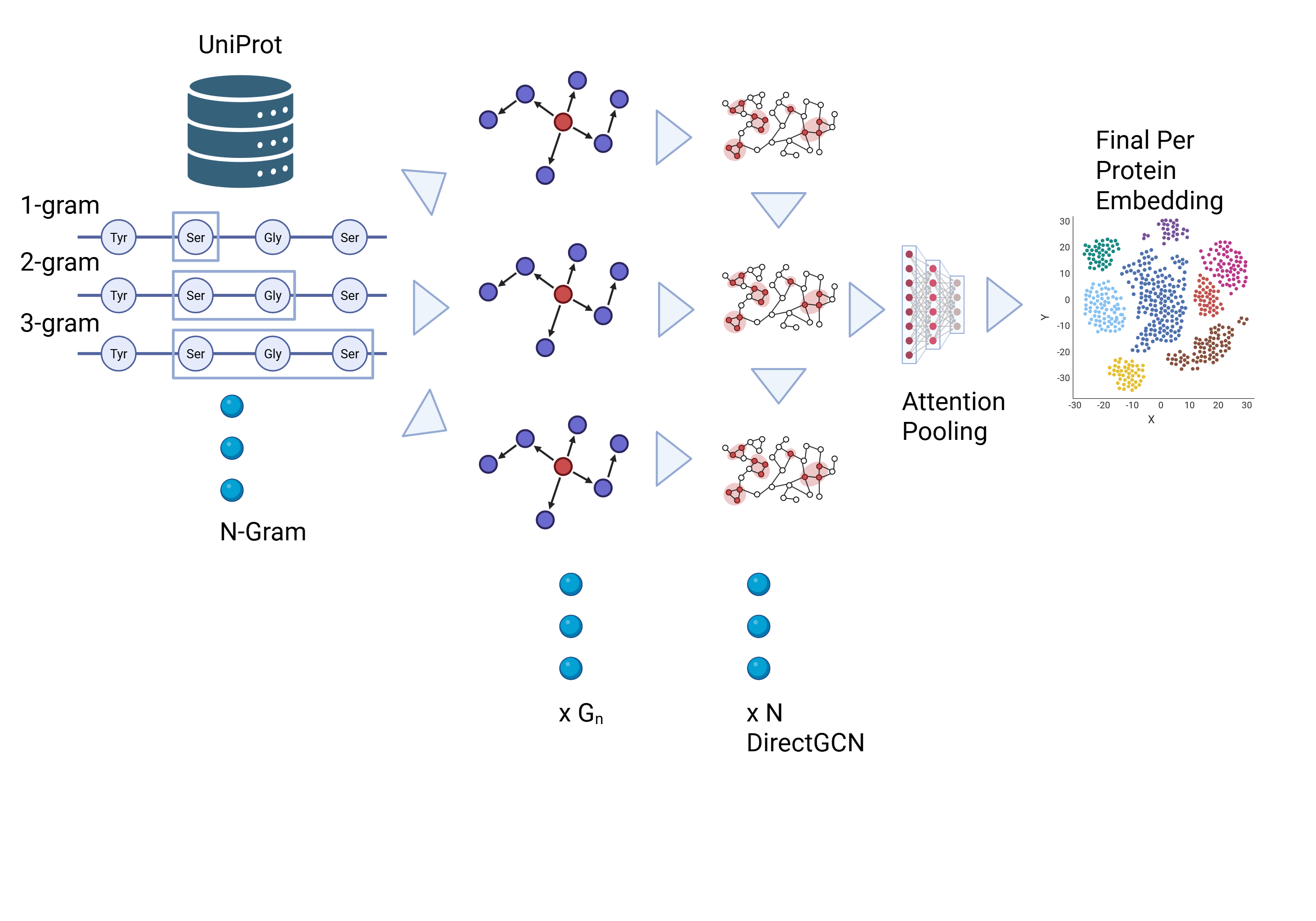} \\
			\textbf{(a)} $ProtGram$: our unique probabilistic approach in modeling protein sequences. \\
			\vspace{1em}
			\begin{tabular}{cc}
				\includegraphics[width=0.48\linewidth]{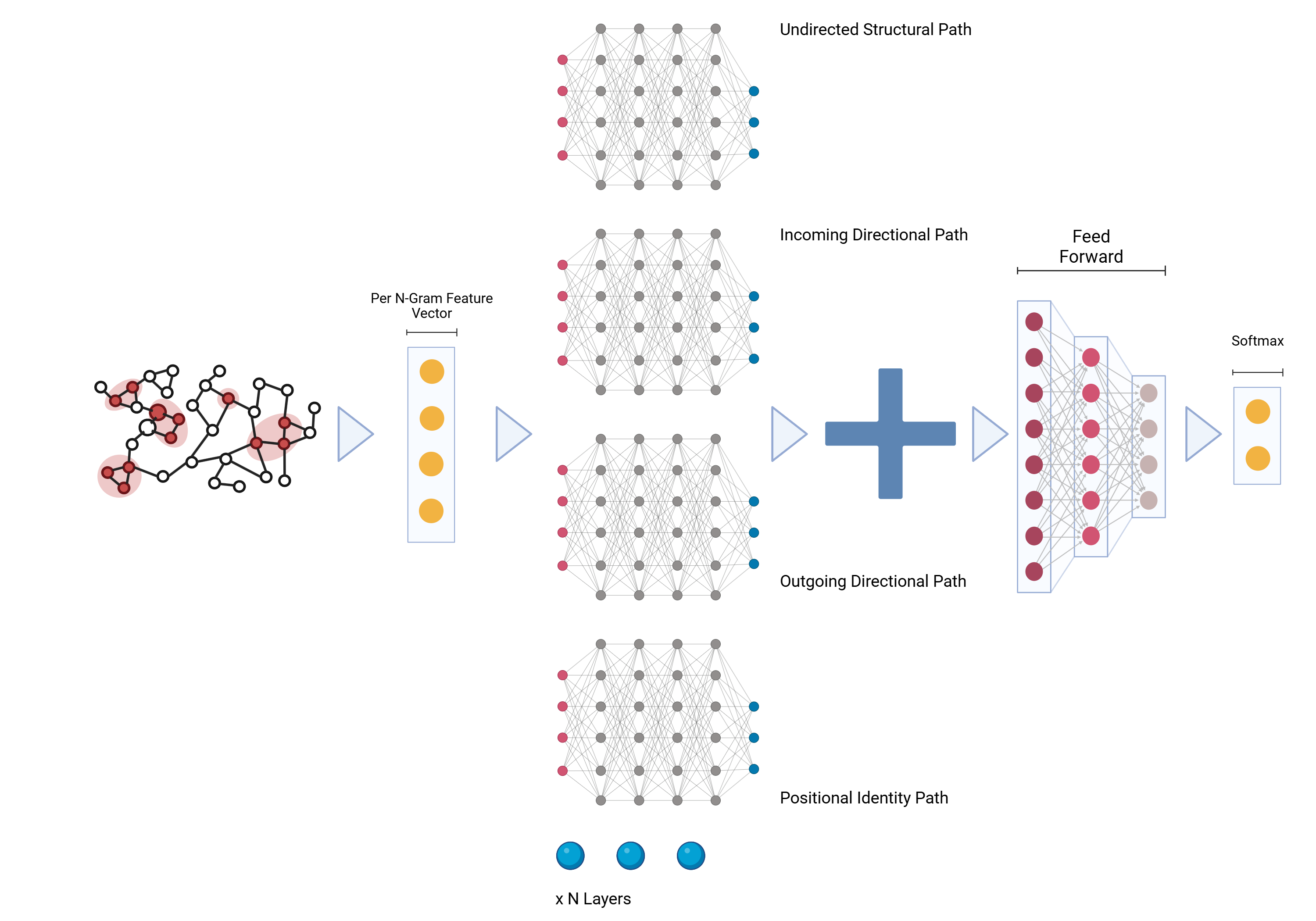} &   
				\includegraphics[width=0.48\linewidth]{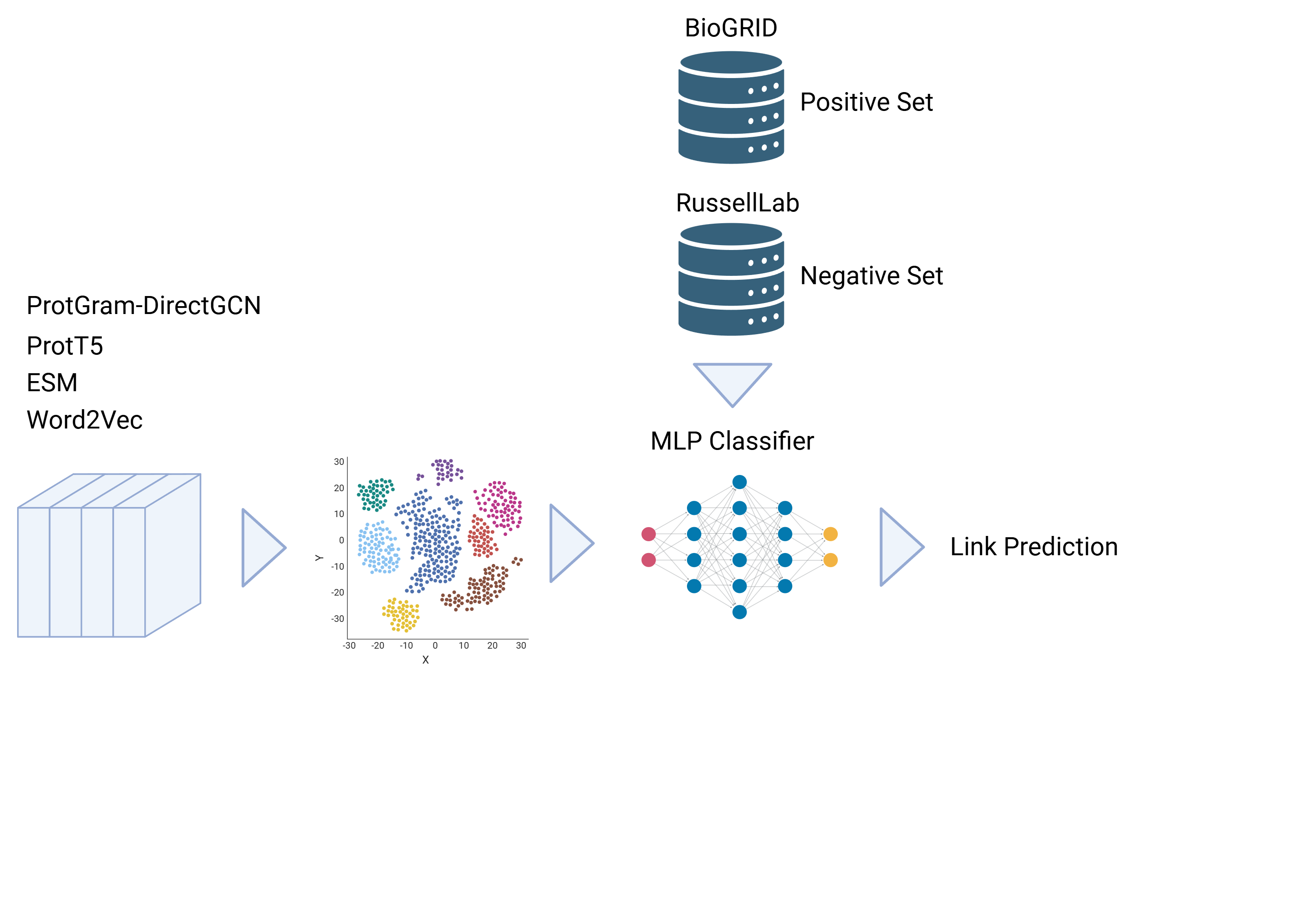} \\
				\begin{minipage}[t]{0.48\linewidth}
					\centering
					\textbf{(b)} $DirectGCN$: the layer is composed of multiple paths extracted from the graph; unstructured, directional in addition to a positional identity. The paths are then aggregated and transformed via a final feed forward layer and is trained on a next node prediction classification task.
				\end{minipage}                                                               &   
				\begin{minipage}[t]{0.48\linewidth}
					\centering
					\textbf{(c)} $PPI$: the final embeddings are pooled via an attention pooling layer then passed down to the downstream task of classification based link prediction where a classifier is trained on a standard ground truth data.
				\end{minipage} \\
			\end{tabular}
		\end{tabular}
		\caption{$ProtGram-DirectGCN$ full pipeline.}
		\label{fig:model}
	\end{figure*}

	The full $DirectGCN$ model is composed of a stack of these custom hybrid layers. The overall architecture is defined as follows:
	\begin{itemize}
		\item \textbf{Input Layer}: The initial node features for the n-grams, $H^{(0)} \in \mathbb{R}^{N \times F^{(0)}}$, are either identity initialized (for $n=1$) or derived from the embeddings of the previous n-gram level (for $n>1$).
		\item \textbf{Hidden Layers}: The model stacks $L$ hidden layers. For each layer $l \in \{0, \dots, L-1\}$, the output $H^{(l+1)}$ is computed by applying the $DirectGCN$ layer transformation (Equations 3-9) to the previous layer's output $H^{(l)}$. A residual connection is included to facilitate deeper architectures, and a Leaky ReLU activation function followed by dropout is applied after each layer to introduce non-linearity and prevent overfitting.
		\item \textbf{Output Layer}: The output of the final hidden layer, $H^{(L)}$, serves as the learned n-gram embeddings, $Z_{n-gram}$ for the auxiliary node classification tasks (community detection \cite{blondel_fastunfoldingcommunities_2008} or next node prediction) on $G_{n}$. These embeddings are passed through a final linear decoder, which is a small feed forward layer, to produce the final class prediction logits:
		\begin{equation}
			\text{Logits} = \text{Decoder}(Z_{n-gram})
		\end{equation}
		A $LogSoftmax$ function is then applied to these logits for training with a negative log-likelihood loss. The final embeddings, $Z_{n-gram}$, are L2-normalized before extraction. Please refer to figure \ref{fig:model} for a complete overview of $ProtGram-DirectGCN$.
	\end{itemize}
	
	\section{Experiments}
	This section outlines the experimental design employed to evaluate the proposed $ProtGram-DirectGCN$ model. Our experiments are structured to: (1) assess the intrinsic performance of $DirectGCN$ on standard graph benchmark datasets; (2) detail the construction of a hierarchy of global n-gram graphs $G_{n}$ from the UniProt sequence dataset a method we call $ProtGram$; (3) evaluate the ability of $ProtGram-DirectGCN$ to learn meaningful representations from these $G_{n}$ graphs via a self-supervised pre-training task; (4) apply these learned representations, after pooling to the protein level, to the downstream task of $PPI$ link prediction; and (5) compare the efficacy of our protein embeddings with those derived from the state-of-the-art PLMs and other baselines.
	
	\subsection{Materials}
	Our approach in this investigation is to minimize any cleaning or modification of publicly available datasets before running our main computational pipeline. UniProt-SPROT (current state) and UniRef50 (future work) are our main sources of protein sequences. All sequences are cleaned only after an official, validated, automated download. In this context, "prior processing" refers specifically to any data cleaning or manipulation done before the computational pipeline starts. In contrast, "preprocessing" refers to steps such as tokenization and the removal of special characters from sequences, which occur immediately before the main pipeline and as part of it. Sequence cleaning is dynamic; when a pipeline component requests sequence data, the raw FASTA file is read and any character not representing one of the 20 standard amino acids (A, C, D, E, F, G, H, I, K, L, M, N, P, Q, R, S, T, V, W, Y), in addition to an added separator token between individual sequences, is filtered out. This standardizes the alphabet for all downstream models. Sequence lengths are capped at $10,000$ characters to fit in memory and on the GPU during training. Tokenization, or preprocessing, occurs on the fly for all models tested, including our own. For PPI ground truth we apply automated preprocessing to positive and negative ground truth links. Positive links are automatically downloaded from BioGRID \cite{oughtred_biogriddatabasecomprehensive_2021}, and negative links are obtained from \cite{trabuco_negativeproteinprotein_2012}. Identifiers from these raw datasets are converted to canonical UniProtKB IDs using a mapping database built directly from the official UniProt ID mapping database. The standardized interaction pairs are saved in Parquet format and serve as the definitive ground truth for all evaluation tasks. Further preparation details are in the supplementary material.
	
	\subsubsection{Construction of Hierarchical $G_{n}$ Graphs via $ProtGram$}
	The primary dataset for our methodology is a hierarchy of global n-gram graphs, $G_{n}$, constructed from the UniProt Swiss-Prot sequence database.
	\begin{itemize}
		\item \textbf{Corpus}: We used the curated and reviewed UniProt Swiss-Prot dataset, containing $573,230$ protein sequences. Larger and more diverse sequence files liek UniRef50 and UniRef100 and PDB are in our plans to train our model on.
		\item \textbf{Graph Construction}: For each n-gram level from $n=1$ to $n=3$, we constructed a separate graph $G_{n}$. The nodes $V_{n}$ are the unique n-grams of length $n$ found in the corpus. A directed edge $(u, v)$ exists if n-gram $v$ can be formed by shifting a one-character window over n-gram $u$. The edge weight $w_{uv}$ is the total frequency of this transition across all sequences. This process resulted in $3$ graphs of increasing size and complexity, as detailed in table \ref{tab:gn_stats}.
	\end{itemize}
	
	\begin{table}[h!]
		\centering
		\caption{Statistics of the constructed n-gram graphs ($G_{n}$).}
		\label{tab:gn_stats}
		\begin{tabular}{crr}
			\toprule
			\textbf{n-gram Level (n)} & \textbf{\# Nodes (Unique n-grams)} & \textbf{\# Edges (Unique Transitions)} \\
			\midrule
			1                         & 21                                 & 601                                    \\
			2                         & 601                                & 10,669                                 \\
			3                         & 10,669                             & 180,273                                \\
			4                         & 180,273                            & 3,240,330                              \\
			\bottomrule
		\end{tabular}
	\end{table}
	
	\subsection{Intrinsic Evaluation of $DirectGCN$}
	To establish the general graph representation learning capabilities of the  $DirectGCN$ architecture, we first evaluated it on commonly used public benchmark datasets for node classification. Though it  is very important to note that our custom model is designed specifically for the type of hierarchical n-gram graph inferred from protein sequences so the goal is not to evaluate how superior our model is to other standard GNNs but rather to establish validity of the capability of our model to process graph data. For example you will see in the results that our model might not be the best at processing highly homophilic sparse non-directional graphs like Citeseer and Cora.
	
	\begin{itemize}
		\item \textbf{Datasets}: We selected standard GNN benchmark datasets for the intrinsic evaluation: Karate Club, Cora, CiteSeer, PubMed, Cornell, Texas, and Wisconsin. All of the datasets where downloaded from the official \href{https://pytorch-geometric.readthedocs.io/en/2.6.0/modules/datasets.html}{PyTorch Geometric repository}. For each dataset, we evaluated performance on their original edges (potentially directed). See table \ref{tab:dataset_stats}.
		
		\begin{table}[h!]
			\centering
			\caption{Statistics of standard datasets used in the benchmark evaluation.}
			\label{tab:dataset_stats}
			\begin{tabular}{lrrrr}
				\toprule
				\textbf{Dataset} & \textbf{\# Nodes} & \textbf{\# Edges} & \textbf{\# Features} & \textbf{\# Classes} \\ \midrule
				Karate Club      & 34             	 & 78            	 & 0                
				& 4 				  \\
				Cora             & 2,708             & 10,556            & 1,433                & 7                   \\
				PubMed           & 19,717            & 88,648            & 500                  & 3                   \\
				Cornell          & 183               & 298               & 1,703                & 5                   \\
				Texas            & 183               & 325               & 1,703                & 5                   \\
				Wisconsin        & 251               & 515               & 1,703                & 5                   \\ \bottomrule
			\end{tabular}
		\end{table}
		
		\item \textbf{Task \& Setup}: The task was semi-supervised node classification relying on a fixed 10\%/10\%/80\% train/validation/test split. All models were trained for 300 epochs using the Adam optimizer with a fixed 2 layer and layer norm architecture.
		
		\item \textbf{Baseline Models}: Graph Convolutional Network (GCN) \cite{kipf_semisupervisedclassificationgraph_2017}, Graph Attention Network (GAT) \cite{velickovic_graphattentionnetworks_2018}, GraphSAGE \cite{hamilton_inductiverepresentationlearning_2018}, Graph Isomorphism Network (GIN) \cite{xu_howpowerfulare_2019}, and DirGNN \cite{rossi_edgedirectionalityimproves_2023}.
		
		\item \textbf{Results}: The goal of this evaluation was to validate $ProtGram-DirectGCN$ as a sound GNN architecture. On high-homophily citation networks (Cora, CiteSeer, PubMed), $ProtGram-DirectGCN$ underperformed relative to simpler models like GCN and GAT. This is an expected outcome, as its complex, over-parameterized architecture is not well-suited for these tasks and struggles to converge effectively. However, on the heterophilic WebKB datasets (Cornell, Texas, Wisconsin), where relationships are more complex, its performance was more reflective of its innate capacity. This validates that the model is functional but highly specialized, justifying its application to our custom, heterophilic n-gram graphs rather than general-purpose benchmarks. A summary of results is presented in table \ref{tab:benchmark_performance}.
	\end{itemize}
	
	\begin{longtable}{llrrrr}
		\caption{Model performance on directed datasets. Accuracy and F1-Score are reported as $mean \pm std$. (M) denotes macro average. Bold indicates best performance.}
		\label{tab:benchmark_performance}\\
		\toprule
		Dataset   & Model     & Accuracy                     & F1-Score (M)                 & Precision (M)   & Recall (M)      \\
		\midrule
		\endfirsthead
		\caption[]{-- continued from previous page}\\
		\toprule
		Dataset   & Model     & Accuracy                     & F1-Score (M)                 & Precision (M)   & Recall (M)      \\
		\midrule
		\endhead
		\midrule
		\multicolumn{6}{r}{{Continued on next page...}} \\
		\endfoot
		\bottomrule
		\endlastfoot
		
		& GCN       & 0.8722 $\pm$ 0.0088          & 0.8622 $\pm$ 0.0106          & 0.8629          & 0.8651          \\
		& GAT       & 0.8863 $\pm$ 0.0062          & 0.8754 $\pm$ 0.0099          & 0.8808          & 0.8723          \\
		Cora      & GIN       & 0.8671 $\pm$ 0.0103          & 0.8588 $\pm$ 0.0134          & 0.8637          & 0.8575          \\
		& DirectGCN & 0.8590 $\pm$ 0.0189          & 0.8480 $\pm$ 0.0256          & 0.8493          & 0.8497          \\
		& DirGNN    & 0.8530 $\pm$ 0.0142          & 0.8407 $\pm$ 0.0172          & 0.8449          & 0.8400          \\
		\midrule
		& GCN       & 0.8631 $\pm$ 0.0047          & 0.8553 $\pm$ 0.0056          & 0.8573          & 0.8542          \\
		& GAT       & 0.8529 $\pm$ 0.0089          & 0.8456 $\pm$ 0.0103          & 0.8491          & 0.8445          \\
		PubMed    & GIN       & 0.8716 $\pm$ 0.0052          & 0.8669 $\pm$ 0.0054          & 0.8652          & 0.8695          \\
		& DirectGCN & 0.8451 $\pm$ 0.0053          & 0.8370 $\pm$ 0.0067          & 0.8360          & 0.8392          \\
		& DirGNN    & 0.8107 $\pm$ 0.0120          & 0.8000 $\pm$ 0.0116          & 0.8022          & 0.8007          \\
		\midrule
		& GCN       & 0.4101 $\pm$ 0.0608          & 0.2440 $\pm$ 0.0590          & 0.2406          & 0.2713          \\
		& GAT       & 0.4264 $\pm$ 0.0748          & 0.1684 $\pm$ 0.0432          & 0.1989          & 0.2202          \\
		Cornell   & GIN       & 0.4862 $\pm$ 0.0770          & 0.3603 $\pm$ 0.0546          & 0.3682          & 0.3974          \\
		& DirectGCN & \textbf{0.5571 $\pm$ 0.0499} & \textbf{0.4104 $\pm$ 0.0837} & \textbf{0.5182} & \textbf{0.4061} \\
		& DirGNN    & 0.5520 $\pm$ 0.0316          & 0.2976 $\pm$ 0.0547          & 0.3096          & 0.3356          \\
		\midrule
		& GCN       & 0.3773 $\pm$ 0.0923          & 0.1640 $\pm$ 0.0403          & 0.1575          & 0.1770          \\
		& GAT       & 0.5464 $\pm$ 0.0567          & 0.2139 $\pm$ 0.0461          & 0.2163          & 0.2569          \\
		Texas     & GIN       & 0.4045 $\pm$ 0.0585          & 0.2199 $\pm$ 0.0369          & 0.2305          & 0.2414          \\
		& DirectGCN & \textbf{0.6940 $\pm$ 0.0202} & \textbf{0.5212 $\pm$ 0.0831} & \textbf{0.6044} & \textbf{0.5071} \\
		& DirGNN    & 0.5353 $\pm$ 0.0540          & 0.2310 $\pm$ 0.0496          & 0.2390          & 0.2703          \\
		\midrule
		& GCN       & 0.4224 $\pm$ 0.0627          & 0.2491 $\pm$ 0.0702          & 0.2600          & 0.2641          \\
		& GAT       & 0.4898 $\pm$ 0.0937          & 0.2413 $\pm$ 0.0686          & 0.3067          & 0.2635          \\
		Wisconsin & GIN       & 0.4219 $\pm$ 0.0625          & 0.2876 $\pm$ 0.0757          & 0.2910          & 0.3000          \\
		& DirectGCN & \textbf{0.6293 $\pm$ 0.0423} & \textbf{0.3833 $\pm$ 0.0584} & \textbf{0.3835} & \textbf{0.4079} \\
		& DirGNN    & 0.4975 $\pm$ 0.1026          & 0.2695 $\pm$ 0.0784          & 0.2934          & 0.2814          \\
	\end{longtable}
	
	\subsection{Learning N-gram Embeddings from $G_{n}$ via Training $ProtGram-DirectGCN$}
	The constructed $G_{n}$ graphs via $ProtGram$ serve as the foundation for learning informative vector representations (embeddings) for each n-gram. This is achieved through a self-supervised training task designed to force the model to understand the sequential grammar inherent in the protein sequences from which the graph was built.
	\begin{itemize}
		\item \textbf{Next-Node Prediction as a Self-Supervised Task}: For each n-gram node $u \in V_n$, we define its label $y_u$ as its most likely successor in the sequence. This successor is determined by identifying the outgoing edge $(u,v)$ with the highest transition frequency (weight) in the raw graph. The task for the GNN is therefore to predict this most probable next n-gram for every node in the graph. This objective compels the model to learn embeddings that encode the sequential and transitional logic of the n-gram language. An n-gram's representation becomes a function of not only its own identity but also the likely sequences it participates in. Final n-gram level is trained on a Louvain community detection \cite{blondel_fastunfoldingcommunities_2008} label task. The community detection is analgous to a larger context window in the graph or a larger neighborhood aggregation. Community detection can be difficult to detect in the smaller n-gram levels because of the inherent faint signal associated with each node but as the number of n-gram levels increase the signal becomes more discriminating of n-gram graph communities. 
		
		\item \textbf{Hierarchical Training}: The training process is hierarchical. For the base level ($n=1$), node features are identity initialized. For each subsequent level $n > 1$, the initial features for a given n-gram node are generated by attention-pooling the final, learned embeddings of its two constituent (n-1)-gram nodes from the previously trained level. This creates a rich, multi-scale representation, where higher-order n-gram features are built upon the learned representations of their sub-components.
		
		\item \textbf{Implementation Details}: The model for each level $n$ is trained for a set number of epochs using the Adam optimizer and a negative log-likelihood loss function on the next-node prediction task. For larger graphs ($n \geq 3$), a Cluster-GCN \cite{chiang_clustergcnefficientalgorithm_2019} approach is used to partition the graph into mini-batches via community detection for memory-efficient training. The final output of this stage is a comprehensive set of learned embeddings for all n-grams at the highest level, $n=3$. Please see the supplementary material for experimental details.
	\end{itemize}
	
	\subsection{Protein-Protein Interaction (PPI) Prediction as Link Prediction}
	\label{subsec:link}
\begin{figure}
		\centering
		\includegraphics[width=0.9\linewidth]{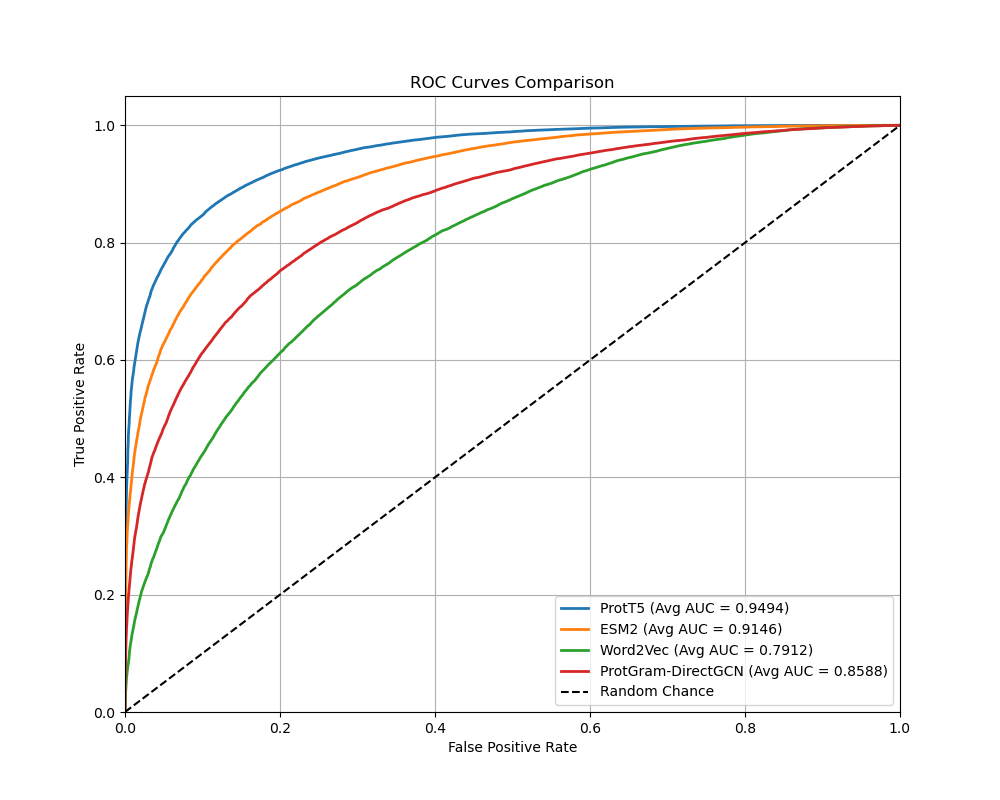}
		\caption{The plot displays the Receiver Operating Characteristic (ROC) curves comparing the performance of protein embeddings generated by the proposed $ProtGram-DirectGCN$ method against the state-of-the-art ProtT5 and ESM language models, in addition to Word2Vec. The evaluation is for the downstream task of Protein-Protein Interaction (PPI) link prediction, with this specific chart illustrating the results from the average of a 5-fold cross-validation. All models perform significantly better than random chance (dashed line). This visualization confirms that while the proposed graph-based method captures a strong predictive signal for protein interactions, both ProtT5 and ESM models serves as a higher-performing benchmark in this experiment.}
		\label{fig:3}
	\end{figure}

\begin{figure}
		\centering
		\includegraphics[width=0.9\linewidth]{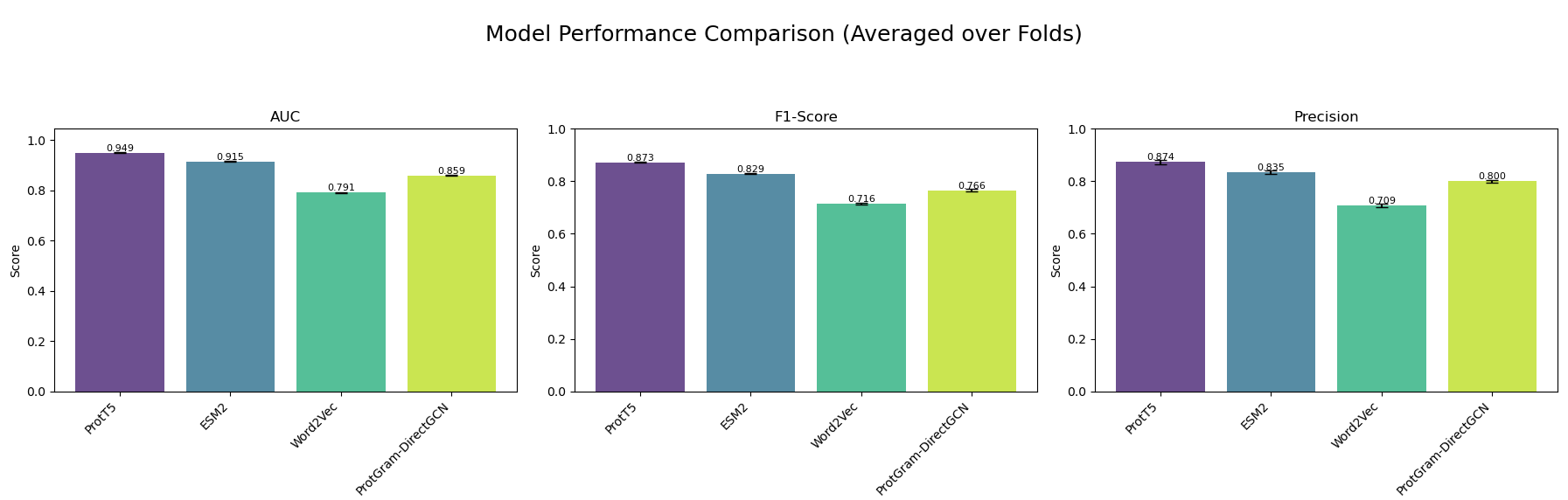}
		\caption{The figure illustrates the models' performance metrics reflecting the results in table \ref{tab:ppi_performance_results}}
		\label{fig:4}
	\end{figure}

	\begin{itemize}
		\item \textbf{Protein-Level Embeddings Generation via Attention Pooling}: A single, fixed-size feature vector is generated for each protein in the UniProt dataset. This is achieved by taking the sequence of each protein, identifying all of its constituent n-grams, retrieving their learned embeddings from the final $ProtGram-DirectGCN$ model, and aggregating these vectors via attention pooling. This results in a single vector that summarizes the global n-gram statistics for each protein. To standardize the feature space for comparison with other methods, Principal Component Analysis (PCA) is applied to reduce the final embeddings dimension to $64$.
		
		In this step, we use self-attention to create a single embedding vector for a protein from its n-gram embeddings (also called residue embeddings). Each n-gram determines its importance within context and receives a unique attention weight, so n-grams matching the protein's syntax have greater influence. This lets the model focus on the most relevant sequence parts. Attention pooling is particularly suited for protein sequences, as it highlights structural motifs that affect binding sites, discussed further in section \ref{subsec:significance}.
		
		We compute attention scores as follows:
		\begin{itemize}
			\item Let the set of n-gram embedding vectors for a protein be $\mathbf{P} = {\mathbf{v}_1, \mathbf{v}_2, ..., \mathbf{v}_n}$, where n is the number of n-grams and each $\mathbf{v}_i$ is a d-dimensional vector.
			
			\item First, we calculate the average of all n-gram embeddings for the protein. This vector, called the context vector, represents the typical pattern or summary of the entire protein sequence. The context vector guides the model in determining which n-grams are most relevant in the protein's context. The context vector $\mathbf{c}$ is the mean of all n-gram vectors: $\mathbf{c} = \frac{1}{n} \sum_{i=1}^{n} \mathbf{v}_i.$
			
			\item Then we score each n-gram vector by its dot product with the context vector $\mathbf{c}$: $s_i = \mathbf{v}_i \cdot \mathbf{c}$. Higher scores indicate greater alignment.
			
			\item The final weights $alpha_i$ come from applying Softmax to the raw scores. This step normalizes scores into a probability distribution: $\alpha_i = \frac{\exp(s_i)}{\sum_{j=1}^{n} \exp(s_j)}$.
			
			\item We next compute the weighted average of the n-gram embeddings using the attention weights. This produces the final, attention-pooled per-protein embedding vector. The final per-protein embeddings is the weighted sum of the n-gram vectors, each scaled by its attention weight: $\mathbf{v}_{\text{protein}} = \sum_{i=1}^{n} \alpha_i \mathbf{v}_i$.
			
			\item This final vector, $\mathbf{v}_{\text{protein}}$, represents the protein, reflecting the contribution of its most important n-grams as determined by self-attention.
		\end{itemize}

		\item \textbf{PPI Datasets}: A benchmark PPI dataset is compiled automatically using known positive interactions from the BioGRID database \cite{oughtred_biogriddatabasecomprehensive_2021} and high-quality negative interactions (non-interacting pairs) from the experimentally-derived Russell Lab datasets \cite{trabuco_negativeproteinprotein_2012}. This ensures a robust and biologically relevant evaluation set.
		
		\item \textbf{Link Prediction Model}: A standard Multi-Layer Perceptron (MLP) was used as the binary classifier. For a pair of proteins $(P_a, P_b)$, the input to the MLP was the concatenation of their embedding vectors.
		
		\item \textbf{Evaluation and Baselines}: The model's performance is rigorously assessed using a 5-fold stratified cross-validation scheme to ensure that results are robust and not dependent on a single random data split. We measure performance using a suite of standard binary classification and ranking metrics, including Area Under the ROC Curve (AUC), F1-Score, Precision, Recall. To contextualize our results, we compare the performance of our $ProtGram-DirectGCN$-derived embeddings against ProtT5 \cite{elnaggar_prottransunderstandinglanguage_2022} available via UniProt. In addition we compare it against ESM \cite{rao_transformerproteinlanguage_2020} where we performed the inference and the embedding generation manually. And finally to further contextualize our work we train a Word2Vec model \cite{mikolov_distributedrepresentationswords_2013} on the concatenated sequences with a context window of 10 tokens with skip-gram and negative sampling for 10 epochs. The exact same MLP architecture and evaluation protocol are used for all embeddings generated to ensure a fair comparison. A hyperparameter optimization protocol was applied to find the best parameters for the MLP model. See table \ref{tab:ppi_performance_results}, figure \ref{fig:3} and figure \ref{fig:4} for the final evaluation of all models.
	\end{itemize}
	
	\begin{table*}[htbp]
		\centering
		\caption{Performance Comparison of Protein Embeddings on PPI Link Prediction (Averaged over 5 Folds)}
		\label{tab:ppi_performance_results}
		\resizebox{\textwidth}{!}{
			\begin{tabular}{lcccc}
				\toprule
				\textbf{Embedding Method} & \textbf{AUC}      & \textbf{F1}       & \textbf{Precision} & \textbf{Recall} \\
				\midrule
				ProtT5                    & 0.9494$\pm$0.0011 & 0.8727$\pm$0.0019 & 0.8736             & 0.8720          \\
				ESM                  & 0.9146$\pm$0.0006 & 0.8293$\pm$0.0019 & 0.8351             & 0.8238          \\
				\textit{\textbf{ProtGram-DirectGCN}}         & \textit{\textbf{0.8588}}$\pm$0.0014 & \textit{\textbf{0.7659}}$\pm$0.0049 & \textit{\textbf{0.7998}}             & \textit{\textbf{0.7349}}          \\
				Word2Vec                  & 0.7912$\pm$0.0017 & 0.7159$\pm$0.0029 & 0.7085             & 0.7236          \\
				\bottomrule
			\end{tabular}
		}
	\end{table*}
	
	\subsection{Ablation Study of $ProtGram-DirectGCN$}
	Here we are going to understand the properties of the n-gram graph and its generated residue representation by training the model on a subset of the available sequence data on different model configuration. This step is crucial as the smallest pertubation in the data or the model affect the final per protein embeddings due to the hierarchical nature of the model.
	
	\begin{figure*}[t!]
		\centering
		\begin{tabular}{c}
			\begin{tabular}{cc}
				\includegraphics[width=0.48\linewidth]{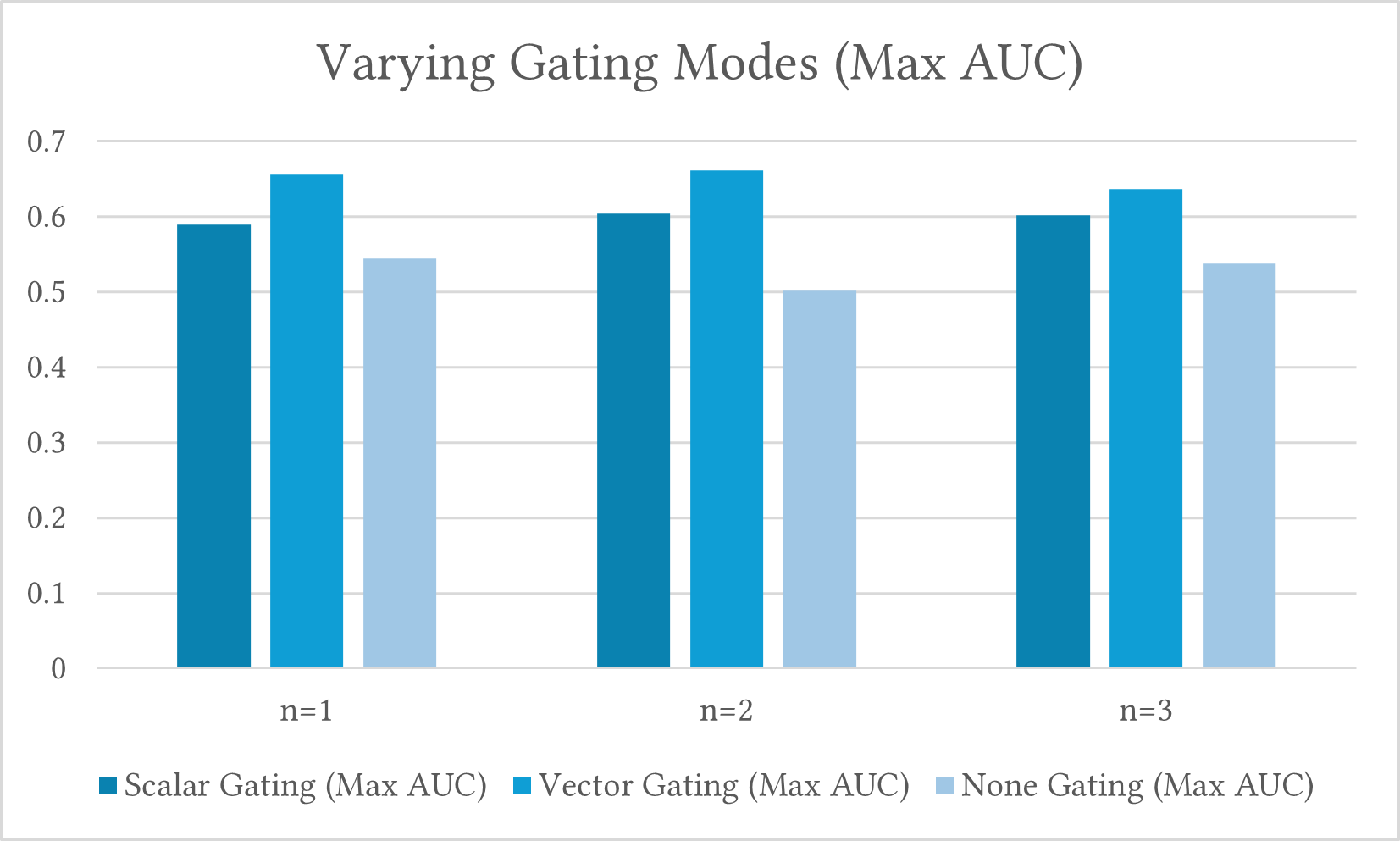} &   
				\includegraphics[width=0.48\linewidth]{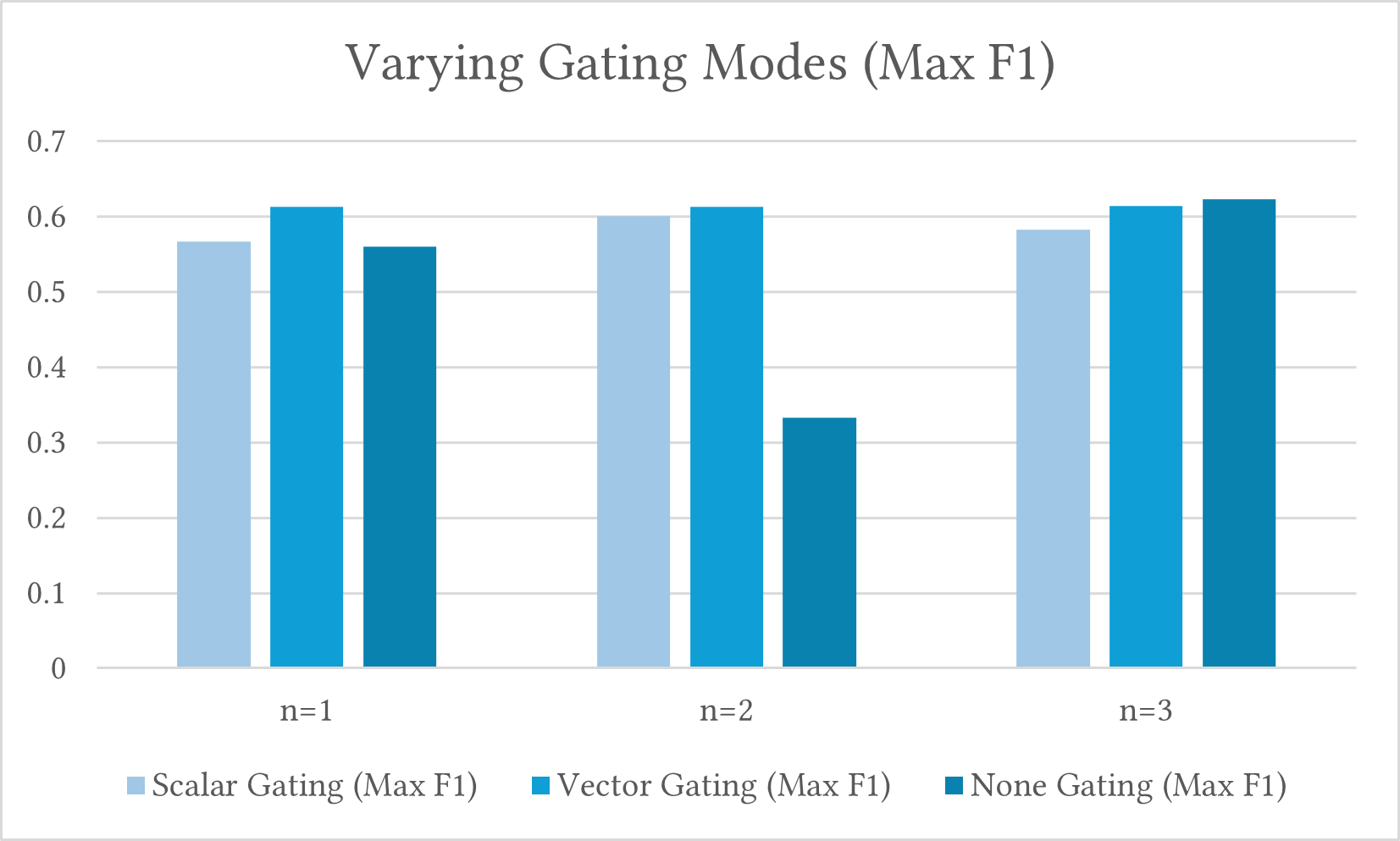} \\
				\begin{minipage}[t]{0.48\linewidth}
					\centering
					\textbf{(b)} This plot backs up our findings. For scalar gating, the AUC peaks at $n=3$ in some summaries but is optimal at $n=2$ in others, with $n=3$ experiencing a decline. This shows that the optimal n-gram depends on experimental factors. For vector gating, performance improves from $n=1$ to $n=2$, then drops at $n=3$. This supports the idea that $n=2$ is optimal, as higher complexity may compromise results. The plot illustrates variation in AUC across different runs, highlighting the model's sensitivity to test conditions.
				\end{minipage}                                                               &   
				\begin{minipage}[t]{0.48\linewidth}
					\centering
					\textbf{(c)} This plot shows that vector gating is the most effective approach, typically yielding the highest F1 scores across all n-gram levels. Scalar gating is generally preferable to no gating. No Gating models have poor results, except for a rare F1 outlier likely due to experimental effects. Vector gating performs best at $n=2$. Scalar gating may peak at n = 2 or n = 3.
				\end{minipage} \\
			\end{tabular}
		\end{tabular}
		\caption{Ablation results for varying gating modes versus different n-gram levels when training $ProtGram-DirectGCN$}
		\label{fig:three_images}
	\end{figure*}
	
	The $ProtGram-DirectGCN$ model uses a hierarchical approach. Embeddings from lower-order n-grams initialize features for higher-order n-gram graphs. Increasing n allows the model to capture richer protein sequence context. The ablation phase used just $5\%$ of UniProt-Sprot to test data augmentation and prediction capabilities in low-data settings we varied two key model components. First, we changed the size of the hierarchical graph; specifically, the value of n in the n-gram. Second, we studied the effect of the gating mechanism in 3 configurations: vector-based, where each graph node has a single gating score; scalar, where each path of the DirectGCN 3-paths has one gating score; and no gating at all. We present the results of this step in figure \ref{fig:three_images} and conclude from our study the following.
	
	\begin{itemize}
		\item{No Gating}: The lack of gating has shown a consistent reduction in predictive power, with AUC scores closer to $50\%$. This suggests a substantial drop in classification ability. In the context of limited and less diverse data, the model struggles to overcome the weaker signal. These results indicate that even in sparse data settings, including some form of gating provides significant benefit, regardless of the model's complexity. 
		
		\item{Scalar Gating}: With scalar gating, AUC modestly increases with n. Notably, $n=3$ shows a statistically significant improvement over $n=1$ and $n=2$, suggesting that 3-grams are more helpful in some cases. While the AUC rises with n, n1 (0.5521), n2 (0.5902), and n3 (0.6016), we occasionally observed diminishing returns or negative impacts when increasing to 3-grams. This suggests that, under certain conditions, 3-grams can introduce noise, lead to overfitting, or result in overly specific features. The shift from 2-grams to 3-grams, therefore, does not universally strengthen performance, emphasizing the need for careful tuning.
		
		\item{Vector Gating}: For vector gating, the AUC increases from n1 (0.6376) to n2 (0.6616), then drops at n3 (0.6367). F1 shows a similar trend. Increasing n from 1 to 2 brings improvement, but going to 3 does not consistently help. This underlines diminishing returns for higher-order n-grams in low-data settings. The transition from 1-grams to 2-grams often enhances both gating types. Moving from 2-grams to 3-grams, however, can result in a decrease in performance. The best n-gram level varies by task or dataset, and higher n may require adjustments to the model or training.
	\end{itemize}
	
	Overall, vector gating consistently outperforms scalar gating across all n-gram levels and test settings. Here, "vector gating" refers to the use of node-specific, learnable gating vectors (the coefficients $C_{*}(l)$ in Equation 7) that allow each node in the graph to control how much information it integrates from each of its neighbors within the DirectGCN layers. This process is analogous to combining multiple features at each node. These coefficients enable the model to adjust the influence of each neighbor at each node, yielding robust and accurate vector representations of nodes. Collectively, our findings indicate that incorporating higher-order n-grams (capped at 2) generally increases diversity in the information processed and strengthens the signal, underscoring the importance of considering the order of residues, or sequence context, in the model. However, increasing n to 3 can bring diminishing or negative returns. Notably, vector gating remains a better mode, highlighting the importance of learnable, node-wise gating in effectively integrating information in complex protein n-gram graphs.
	
	Altogether, these findings demonstrate that $ProtGram-DirectGCN$’s performance is sensitive to both n-gram level and gating mode. Variations in experimental factors can produce notable differences. This reinforces the need for careful model tuning and thorough evaluation before deploying the model. In summary, gating proves crucial. Simply increasing n beyond an optimal point does not always lead to improved predictive power.
	
	\section{Discussion}
	This study introduced and evaluated a novel $ProtGramDirectGCN$ model for learning representations from a globally constructed, directed, dense, and weighted graph of amino acid residues $G_{n}$ derived from the UniProt dataset. The primary objective was to assess the efficacy of this approach for generating informative protein embeddings applicable to downstream biological prediction tasks, particularly Protein-Protein Interaction (PPI) link prediction. This section discusses the main findings, their implications, the limitations of the current work, and promising avenues for future research.
	
	\subsection{Summary of Findings}
	Our experimental evaluations spanned several stages: validating the core $ProtGram-DirectGCN$ architecture on standard GNN benchmarks and applying the derived protein-level embeddings to predict PPIs, including a comparison against state-of-the-art PLMs (ProtT5) and (ESM) embeddings and a standard base line Word2Vec.
	
	\begin{itemize}
		\item \textbf{Evaluation of $DirectGCN$}: The benchmark results in table \ref{tab:benchmark_performance} confirm that $DirectGCN$ is a functionally sound GNN. Its underperformance on high-homophily citation networks and competitive performance on more complex, heterophilic graphs highlights its specialization. The architecture is not designed as a general-purpose GCN but as a specialized tool for capturing the complex, directed, and weighted relationships present in our n-gram residue graphs.
		
		\item \textbf{Evaluation of $ProtGram$}: The hierarchical construction of n-gram graphs up to $n=3$ and even further n-gram such as $n=4$ (table \ref{tab:gn_stats}) resulted in a large, complex graph structure. The successful training of our model on this graph hierarchy demonstrates the feasibility of the approach. The key outcome of this stage is the set of high-dimensional embeddings for $10669$ unique 3-grams, which serve as the basis for our protein-level representations.
		
		\item \textbf{Evaluation of $ProtGram-DirectGCN$}: The performance on the downstream PPI link prediction task (table \ref{tab:ppi_performance_results}) is the ultimate measure of our method's utility.
		The results show that the $ProtGram-DirectGCN$ was able to learn structural features from the protein sequences with reliable discriminating ability and hence was able to demonstrate excellent predictive power on the task with an AUC value above $85\%$ despite being trained on limited sequence data compared to its counter PLMs. The F1 score also demonstrates the model's precision even with a lowered recall and hence missing more positive samples due to the limited capacity of the model. This highly suggested that the construction and inference of the underlying directed graph of amino acid transitions in a hierarchical fashion captures structural and relational features across multiple proteins.
		
		\item \textbf{Comparison with ProtT5, ESM and Word2Vec Embeddings}: The comparison with ProtT5 and ESM embeddings which are generted by the very powerful high capacity T5 encoder-decoder transformer model that is trained on the more comprehensive UniRef50 dataset is not meant as a head-to-head comparison. But rather as a demonstration that hefty transformer architectures for specialized tasks like PPI prediction can be contended with models that capture the underlying dynamics without having to rely on long context windows and demanding computational resources needs. The long range dependencies captured by ProtT5 and ESM are the reason why it is an efficient feature extractor for proteins. Yet those same dynamics can be captured from a lower level faint signal such as the simple transition directed graph of amino acids without long context windows. ProtT5, ESM and $ProtGram-DirectGCN$ rely on computationally expensive preprocessing yet with $ProtGram-DirectGCN$ the significant decrease in the cost of model training especially when the technique gets more established and developed will result in a paradigm shift when it comes to how we think about specialized tasks for LLMs in general. Training a Word2Vec model is usually considered a sanity check as it represents the base line that any predicitve deep learning model should be able to outperform. Including Word2Vec and ESM helps contextualize our work and shows that the results are consistent with what has been reported before in the literature for all based line models.
	\end{itemize}
	
	\subsection{Biological Significance}
	\label{subsec:significance}
\begin{figure*}[t!]
		\centering
		\begin{tabular}{c}
			\begin{tabular}{cc}
				\includegraphics[width=0.48\linewidth]{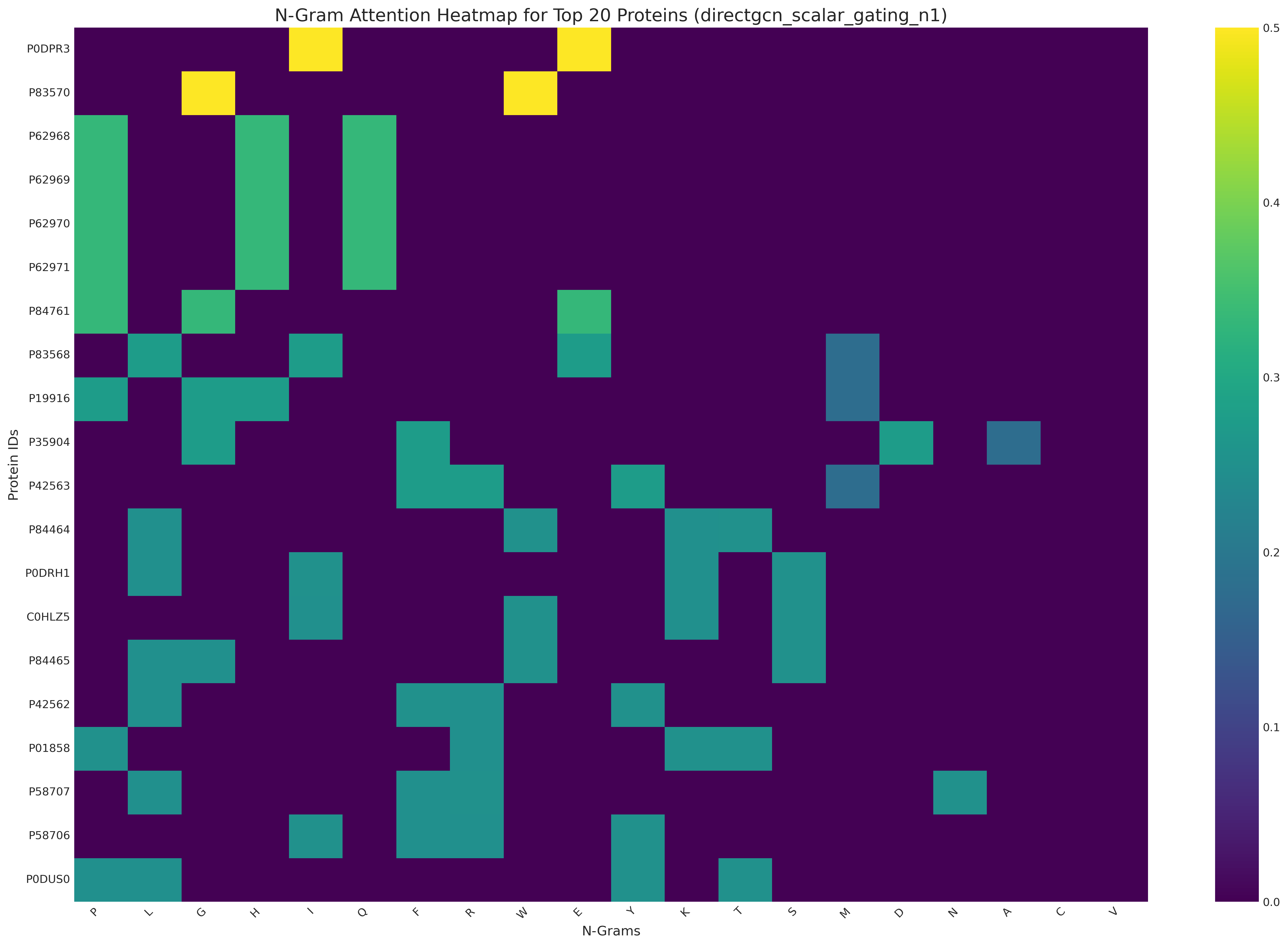} &   
				\includegraphics[width=0.48\linewidth]{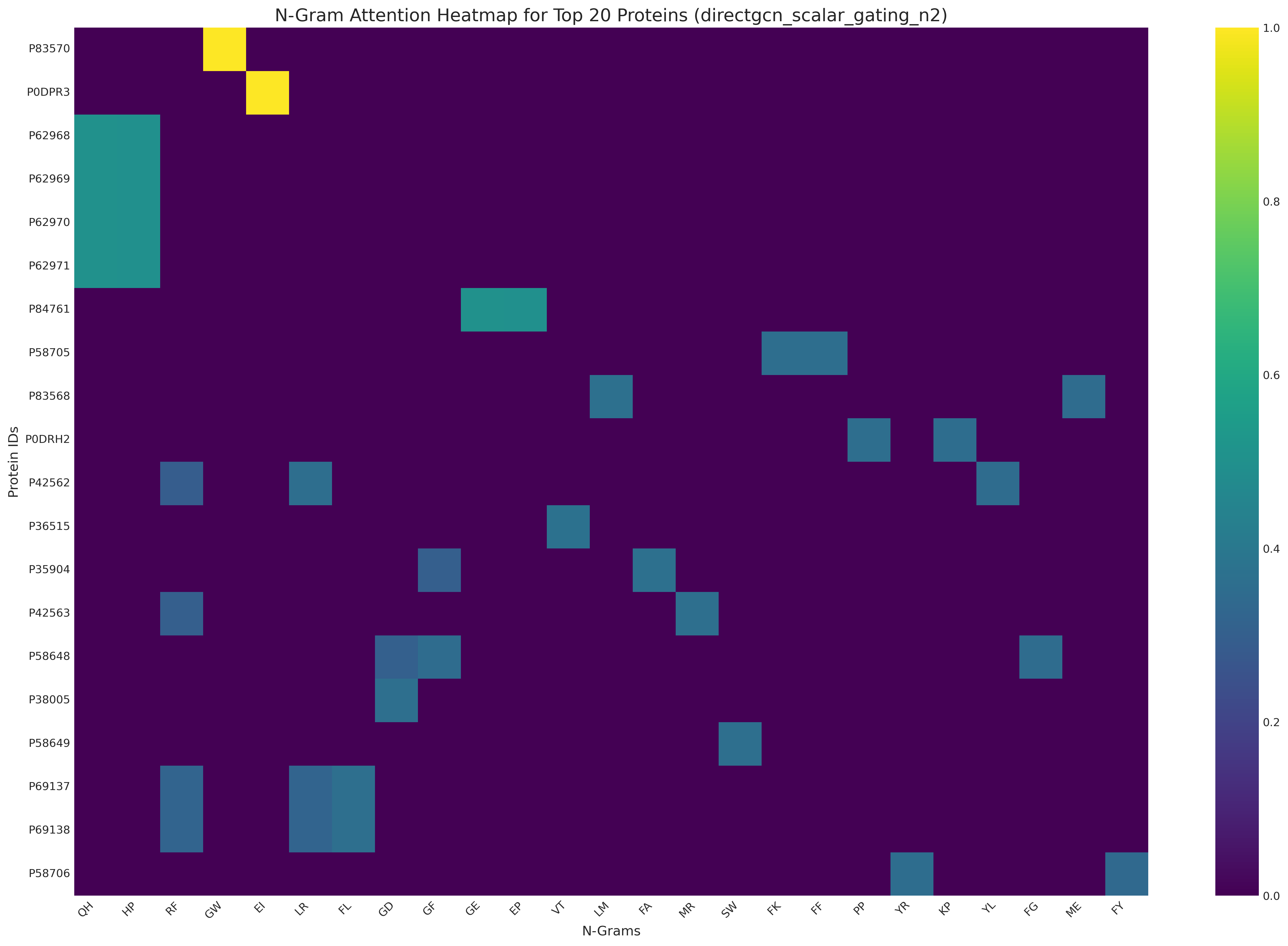} \\
				\begin{minipage}[t]{0.48\linewidth}
					\centering
					\textbf{(b)} $n=1$.
				\end{minipage}                                                               &   
				\begin{minipage}[t]{0.48\linewidth}
					\centering
					\textbf{(c)} $n=2$.
				\end{minipage} \\
			\end{tabular}
		\end{tabular}
		\caption{The attention pooling results described in section \ref{subsec:link} highlight attention scores after pooling residue-level (n-gram) embeddings to protein-level embeddings. These heatmaps are generated by identifying n-gram attention weights for proteins with the highest overall variance in attention. The X-axis of such a heatmap represents various n-grams, and the Y-axis represents specific protein IDs, with the color intensity in each cell indicating the attention score assigned to a particular n-gram within a given protein.}
		\label{fig:heatmaps}
	\end{figure*}

	PPI prediction is a bedrock in drug development, understanding drug efficacy, and many other crucial biomedical fields. Figure \ref{fig:heatmaps} displays the computed attention maps at $n=1$ and $n=2$ post pooling. These maps provide insight into the role of distinct n-grams and how they map to functional groups. High attention scores for specific n-grams within a protein indicate that these sequence fragments are considered most relevant or discriminating by the model. This is especially important for forming the overall protein-level representation. This is particularly true for the downstream task of (PPI) prediction. It implies that highly attended n-grams likely correspond to crucial regions within the protein's primary sequence.
	
	The $ProtGram-DirectGCN$ model is based on the intuition that the transition sequence of amino acids, through their side chains or R-groups, determines how a polypeptide chain folds. This folding, in turn, affects interactions with other molecules. Therefore, n-grams with high attention scores in these heatmaps could represent either specific binding sites or key structural motifs. The model learns to prioritize these n-grams as they are critical for determining if and how proteins interact. Key structural motifs are vital for a protein's overall fold. This, in turn, influences its function and interaction capabilities.
	
	The $DirectGCN$ layer itself is designed to process information through multiple, specialized paths. These include incoming, outgoing, and undirected routes. The model combines these paths using a learnable gating mechanism. The attention pooling layer then aggregates these already contextually enriched n-gram embeddings into a single protein embeddings. Thus, the attention scores on the heatmaps reflect not just the local sequence importance. They also indicate contextual and relational significance within the broader n-gram graph and across different interaction types.
	
	By identifying these lead residues and sequence motifs that contribute significantly to the model's predictions, the attention heatmaps can guide hypothesis generation for experimental testing. They can also accelerate the functional annotation of uncharacterized proteins (which we removed in our data preprocessing). Biologists could use these highlighted n-grams to design targeted experiments. For example, they might conduct site-directed mutagenesis to validate their functional role in protein interactions.
	
	One motivation behind $ProtGram-DirectGCN$ is to address the limited context window size in PLMs. By explicitly modeling broad sequence patterns and transition dynamics through n-gram graphs, the model is designed to capture longer-range dependencies that PLMs might miss due to their window size. The attention heatmaps help visualize how the model uses this broader context. They highlight important n-grams that traditional PLMs might overlook because they fall outside their immediate scope.
	
	\subsection{Limitations}
	While this study presents promising results, several limitations should be acknowledged:
	\begin{itemize}
		\item Due to computational constraints, $G_{n}$ was constructed based on UniProt Swiss-Prot standard sequence database. While providing a high-quality reviewed set, more comprehensive and diverse datasets, such as UniRef50/90/100 or the full UniProtKB, could enrich $G_{n}$ at a significant computational cost potentially increasing the predicitve and discriminating power of $ProtGram-DirectGCN$ to be on-par with PLMs.
		\item Initial features for 1-gram nodes in $G_{1}$ were initialized to identity. Including physicochemical properties as initial features could enhance learning and interpretability.
		\item Simple attention pooling was used to generate protein-level embeddings. More advanced pooling mechanisms were not exhaustively explored and might yield improved representations.
		\item The evaluation was centered on PPI link prediction, hence the utility of embeddings for additional tasks remains to be explored.
		\item While the design of $ProtGram-DirectGCN$ is detailed, and we provided some insight via attention map visualization of the role of distinct n-grams and how they map to functional groups, direct interpretation of what specific n-gram relationships contribute most to its performance or downstream PPI predictions currently relies on indirect evaluation through task performance. Deeper interpretability studies are warranted.
		\item The PPI link prediction task relies on the Russell Lab negative dataset \cite{trabuco_negativeproteinprotein_2012}, which, while experimentally grounded, has inherent assumptions and potential biases based on Yeast two-Hybrid limitations. The choice of negative samples can significantly impact the reported performance of PPI prediction.
		\item The current framework primarily relies on sequence-derived information for constructing $G_{n}$ and generating protein embeddings. Direct integration of 3D structural information was not part of this specific study, but it is a key area for future enhancement, as our focus was on building a pipeline that can operate in more challenging conditions such as limited available information and training data.
	\end{itemize}
	
	\subsection{Future Work}
	The findings and limitations of this study open several avenues for future research. First, future iterations will explore constructing $G_{n}$ with richer edge definitions. For example, we could move beyond simple transitions to incorporate longer-range co-occurrences, apply weights from substitution matrices, or use more informative initial node features for amino acids. Also, $ProtGram-DirectGCN$ could be extended by incorporating attention mechanisms within its directional layers. In addition, we could explore more complex architectures with advanced normalization schemes. Another focus will be on introducing and crafting training tasks that are more specific to proteins and their interactions. We also plan to adapt the model to other tasks, such as predicting Gene Ontology labels for individual proteins. This work focuses solely on the limitations of training data and information, which are represented by relying on a limited subset of protein sequences. To address this, we could use predicted contact maps to inform the edges in $G_{n}$ or in peptide-level graphs similar to \cite{sledzieski_dscripttranslatesgenome_2021}, or incorporate residue-level structural features into the initial residue embeddings. Expanding the framework to explicitly model the hierarchical nature of protein organization (residues $\rightarrow$ peptides $\rightarrow$ proteins $\rightarrow$ interactions $\rightarrow$ interaction networks) and exploring second-degree graphs (graphs of interactions) presents a compelling research direction \cite{jeh_simrankmeasurestructuralcontext_2002} as well. Lastly, investing in advanced interpretability techniques will help understand the "black box," and further optimizing the construction of $G_{n}$ and training $ProtGram-DirectGCN$ for even larger sequence datasets will maximize the information captured.
	
	Finally,  its worth mentioning that a notable class of modern PPI prediction methods leverages 3D structural information, either from experimental sources or high-fidelity predictions from models like AlphaFold2 \cite{jumper_highlyaccurateprotein_2021}. These geometric deep learning approaches, such as GearNet and GVP-GNN , have demonstrated state-of-the-art performance by directly encoding the physical and chemical properties of protein surfaces. While these methods are powerful, their applicability is contingent on the availability of accurate structural data. Our work, with $ProtGram-DirectGCN$, intentionally explores a different and complementary direction. We focus exclusively on the protein's primary sequence, aiming to develop a method that is (1) universally applicable to any protein, including those with unknown or poorly predicted structures, and (2) computationally less intensive, as it does not require the computationally expensive step of structure prediction or the storage of large structural files. By constructing a global n-gram graph, our approach seeks to infer higher-order sequence motifs that serve as a proxy for structural and functional information, providing a robust and scalable alternative for large-scale proteome analysis where structural information may be sparse or unavailable. Hence future work could also explore hybrid models that fuse our learned n-gram representations with structural features for proteins where both are available.
	
	\section{Conclusion}
	This paper introduces a novel approach for protein representation learning, which has been shown to enable \textit{in-silico} PPI prediction via a simpler yet expressive learning model. The method focuses on a novel data model that infers hierarchical global n-gram graphs from protein sequences namely $ProtGram$. In these graphs, n-grams, defined as contiguous sequences of n amino acids in proteins, form the nodes, and edges representing relationships between these n-gram sequences. A custom directed graph convolution learning model, $DirectGCN$, is introduced. This model is designed to learn from n-gram graphs with directed edges (edges have direction, indicating the flow from one n-gram to another), heterophily (connections often occur between nodes representing different types of n-grams), and weighted edges (edges have numerical weights that may represent the strength or frequency of the relationship). The model learned distinctive features that capture protein relations, even with limited training data. This offers a valuable and computationally distinct alternative to large-scale PLMs, such as ProtT5 and ESM, under the evaluated conditions. In the future, graph-based representations will be enriched with multi-modal data, including explicit structural information. The scope of application will expand to more biological problems. Ultimately, this work aims to provide a deeper understanding of the molecular interactions that govern life by introducing new methods for analyzing and understanding protein and gene interactions.
	
	\section*{Conflict of Interest Statement}
	The authors declare that the research was conducted without commercial or financial relationships that could create a conflict of interest.
	
	\section*{Author Contributions}
	IAE: developed the preprocessing of protein sequences in addition to designing the directed GCN architecture
	HT: optimized the performance of the GCN model via parallelization, in addition to describing in writing the article
	PG: performed the evaluation of the model and contributed to the writing of the article  
	
	\section*{Funding}
	No funding was involved in this work.

	\section*{Supplemental Data}
	Please see supplementary material for more information.
	
	\section*{Data Availability Statement}
	The datasets and materials used in this study are available publicly at \url{https://www.russelllab.org/negatives/} and \url{https://downloads.thebiogrid.org/BioGRID}. In addition codes for this study can be found at \url{https://github.com/iebeid/ProtGram-DirectGCN/tree/v2}.
	
	\bibliographystyle{Frontiers-Harvard}
	\bibliography{references}

\begin{thebibliography}{35}
\providecommand{\natexlab}[1]{#1}
\expandafter\ifx\csname urlstyle\endcsname\relax
  \providecommand{\doi}[1]{doi:\discretionary{}{}{}#1}\else
  \providecommand{\doi}{doi:\discretionary{}{}{}\begingroup
  \urlstyle{rm}\Url}\fi
\providecommand{\selectlanguage}[1]{\relax}
\providecommand{\bibAnnoteFile}[1]{%
  \IfFileExists{#1}{\begin{quotation}\noindent\textsc{Key:} #1\\
  \textsc{Annotation:}\ \input{#1}\end{quotation}}{}}
\providecommand{\bibAnnote}[2]{%
  \begin{quotation}\noindent\textsc{Key:} #1\\
  \textsc{Annotation:}\ #2\end{quotation}}

\bibitem[{Anfinsen(1973)}]{anfinsen_principlesthatgovern_1973}
Anfinsen, C.~B. (1973).
\newblock Principles that govern the folding of protein chains.
\newblock \emph{Science} 181, 223--230.
\newblock \doi{10.1126/science.181.4096.223}
\bibAnnoteFile{anfinsen_principlesthatgovern_1973}

\bibitem[{Blondel et~al.(2008)Blondel, Guillaume, Lambiotte
  et~al.}]{blondel_fastunfoldingcommunities_2008}
Blondel, V.~D., Guillaume, J.-L., Lambiotte, R., et~al. (2008).
\newblock Fast unfolding of communities in large networks.
\newblock \emph{J. Stat. Mech.} 2008, P10008.
\newblock \doi{10.1088/1742-5468/2008/10/p10008}
\bibAnnoteFile{blondel_fastunfoldingcommunities_2008}

\bibitem[{Brown et~al.(2020)Brown, Mann, Ryder
  et~al.}]{brown_languagemodelsare_2020}
Brown, T.~B., Mann, B., Ryder, N., et~al. (2020).
\newblock Language {{Models}} are {{Few-Shot Learners}}.
\newblock \emph{Adv. Neural Inf. Process. Syst.} 33, 1901.
\newblock \doi{10.48550/arXiv.2005.14165}
\bibAnnoteFile{brown_languagemodelsare_2020}

\bibitem[{Chiang et~al.(2019)Chiang, Liu, Si
  et~al.}]{chiang_clustergcnefficientalgorithm_2019}
Chiang, W.-L., Liu, X., Si, S., et~al. (2019).
\newblock Cluster-{{GCN}}: {{An Efficient Algorithm}} for {{Training Deep}} and
  {{Large Graph Convolutional Networks}}.
\newblock In \emph{Proceedings of the 25th {{ACM SIGKDD International
  Conference}} on {{Knowledge Discovery}} \& {{Data Mining}}} (New York, NY,
  USA: Association for Computing Machinery), {{Kdd}} '19, 257--266.
\newblock \doi{10.1145/3292500.3330925}
\bibAnnoteFile{chiang_clustergcnefficientalgorithm_2019}

\bibitem[{Cho et~al.(2014)Cho, {van Merri{\"e}nboer}, Gulcehre
  et~al.}]{cho_learningphraserepresentations_2014}
Cho, K., {van Merri{\"e}nboer}, B., Gulcehre, C., et~al. (2014).
\newblock Learning {{Phrase Representations}} using {{RNN
  Encoder}}--{{Decoder}} for {{Statistical Machine Translation}}.
\newblock In \emph{Proceedings of the 2014 {{Conference}} on {{Empirical
  Methods}} in {{Natural Language Processing}} ({{EMNLP}})}, eds. A.~Moschitti,
  B.~Pang, and W.~Daelemans (Doha, Qatar: Association for Computational
  Linguistics), 1724--1734.
\newblock \doi{10.3115/v1/D14-1179}
\bibAnnoteFile{cho_learningphraserepresentations_2014}

\bibitem[{Consortium(2023)}]{theuniprotconsortium_uniprotuniversalprotein_2023}
Consortium, T.~U. (2023).
\newblock {{UniProt}}: The {{Universal Protein Knowledgebase}} in 2023.
\newblock \emph{Nucleic Acids Research} 51, D523--d531.
\newblock \doi{10.1093/nar/gkac1052}
\bibAnnoteFile{theuniprotconsortium_uniprotuniversalprotein_2023}

\bibitem[{Devlin et~al.(2019)Devlin, Chang, Lee
  et~al.}]{devlin_bertpretrainingdeep_2019}
Devlin, J., Chang, M.-W., Lee, K., et~al. (2019).
\newblock {{BERT}}: {{Pre-training}} of {{Deep Bidirectional Transformers}} for
  {{Language Understanding}}.
\newblock In \emph{Proceedings of the 2019 {{Conference}} of the {{North
  American Chapter}} of the {{Association}} for {{Computational Linguistics}}:
  {{Human Language Technologies}}, {{Volume}} 1 ({{Long}} and {{Short
  Papers}})} (Minneapolis, Minnesota: Association for Computational
  Linguistics), 4171--4186.
\newblock \doi{10.18653/v1/N19-1423}
\bibAnnoteFile{devlin_bertpretrainingdeep_2019}

\bibitem[{Dill and MacCallum(2012)}]{dill_proteinfoldingproblem50_2012}
Dill, K.~A. and MacCallum, J.~L. (2012).
\newblock The protein-folding problem, 50 years on.
\newblock \emph{science} 338, 1042--1046.
\newblock \doi{10.1126/science.1219021}
\bibAnnoteFile{dill_proteinfoldingproblem50_2012}

\bibitem[{Elnaggar et~al.(2022)Elnaggar, Heinzinger, Dallago
  et~al.}]{elnaggar_prottransunderstandinglanguage_2022}
Elnaggar, A., Heinzinger, M., Dallago, C., et~al. (2022).
\newblock {{ProtTrans}}: {{Toward Understanding}} the {{Language}} of {{Life
  Through Self-Supervised Learning}}.
\newblock \emph{IEEE Trans Pattern Anal Mach Intell} 44, 7112--7127.
\newblock \doi{10.1109/tpami.2021.3095381}
\bibAnnoteFile{elnaggar_prottransunderstandinglanguage_2022}

\bibitem[{Guo et~al.(2008)Guo, Yu, Wen et~al.}]{guo_usingsupportvector_2008}
Guo, Y., Yu, L., Wen, Z., et~al. (2008).
\newblock Using support vector machine combined with auto covariance to predict
  protein-protein interactions from protein2 sequences.
\newblock \emph{Example of classic ML (SVM) and feature discussion} 3, e47--e47
\bibAnnoteFile{guo_usingsupportvector_2008}

\bibitem[{Hamilton et~al.(2018)Hamilton, Ying, and
  Leskovec}]{hamilton_inductiverepresentationlearning_2018}
[Dataset] Hamilton, W.~L., Ying, R., and Leskovec, J. (2018).
\newblock Inductive {{Representation Learning}} on {{Large Graphs}}.
\newblock \doi{10.48550/arXiv.1706.02216}
\bibAnnoteFile{hamilton_inductiverepresentationlearning_2018}

\bibitem[{Hochreiter and
  Schmidhuber(1997)}]{hochreiter_longshorttermmemory_1997}
Hochreiter, S. and Schmidhuber, J. (1997).
\newblock Long {{Short-Term Memory}}.
\newblock \emph{Neural Comput.} 9, 1735--1780.
\newblock \doi{10.1162/neco.1997.9.8.1735}
\bibAnnoteFile{hochreiter_longshorttermmemory_1997}

\bibitem[{Jeh and Widom(2002)}]{jeh_simrankmeasurestructuralcontext_2002}
Jeh, G. and Widom, J. (2002).
\newblock {{SimRank}}: A measure of structural-context similarity.
\newblock In \emph{Proceedings of the Eighth {{ACM SIGKDD}} International
  Conference on {{Knowledge}} Discovery and Data Mining} (New York, NY, USA:
  Association for Computing Machinery), {{Kdd}} '02, 538--543.
\newblock \doi{10.1145/775047.775126}
\bibAnnoteFile{jeh_simrankmeasurestructuralcontext_2002}

\bibitem[{Jha et~al.(2022)Jha, Saha, and
  Singh}]{jha_predictionproteinprotein_2022}
Jha, K., Saha, S., and Singh, H. (2022).
\newblock Prediction of protein--protein interaction using graph neural
  networks.
\newblock \emph{Sci Rep} 12, 8360.
\newblock \doi{10.1038/s41598-022-12201-9}
\bibAnnoteFile{jha_predictionproteinprotein_2022}

\bibitem[{Jumper et~al.(2021)Jumper, Evans, Pritzel
  et~al.}]{jumper_highlyaccurateprotein_2021}
Jumper, J., Evans, R., Pritzel, A., et~al. (2021).
\newblock Highly accurate protein structure prediction with {{AlphaFold}}.
\newblock \emph{Nature} 596, 583--589.
\newblock \doi{10.1038/s41586-021-03819-2}
\bibAnnoteFile{jumper_highlyaccurateprotein_2021}

\bibitem[{Kipf and Welling(2017)}]{kipf_semisupervisedclassificationgraph_2017}
Kipf, T.~N. and Welling, M. (2017).
\newblock Semi-{{Supervised Classification}} with {{Graph Convolutional
  Networks}}.
\newblock \emph{arXiv:1609.02907 [cs.LG]} \doi{10.48550/arXiv.1609.02907}
\bibAnnoteFile{kipf_semisupervisedclassificationgraph_2017}

\bibitem[{Lee et~al.(2020)Lee, Yoon, Kim
  et~al.}]{lee_biobertpretrainedbiomedical_2020}
Lee, J., Yoon, W., Kim, S., et~al. (2020).
\newblock {{BioBERT}}: A pre-trained biomedical language representation model
  for biomedical text mining.
\newblock \emph{Bioinformatics} 36, 1234--1240.
\newblock \doi{10.1093/bioinformatics/btz682}
\bibAnnoteFile{lee_biobertpretrainedbiomedical_2020}

\bibitem[{Mikolov et~al.(2013)Mikolov, Sutskever, Chen
  et~al.}]{mikolov_distributedrepresentationswords_2013}
Mikolov, T., Sutskever, I., Chen, K., et~al. (2013).
\newblock Distributed {{Representations}} of {{Words}} and {{Phrases}} and
  their {{Compositionality}}.
\newblock In \emph{Advances in {{Neural Information Processing Systems}}}
  (Curran Associates, Inc.), vol.~26
\bibAnnoteFile{mikolov_distributedrepresentationswords_2013}

\bibitem[{Oughtred et~al.(2021)Oughtred, Rust, Chang
  et~al.}]{oughtred_biogriddatabasecomprehensive_2021}
Oughtred, R., Rust, J., Chang, C., et~al. (2021).
\newblock The {{BioGRID}} database: {{A}} comprehensive biomedical resource of
  curated protein, genetic, and chemical interactions.
\newblock \emph{Protein Science} 30, 187--200.
\newblock \doi{10.1002/pro.3978}
\bibAnnoteFile{oughtred_biogriddatabasecomprehensive_2021}

\bibitem[{Perkins et~al.(2010)Perkins, Diboun, Dessailly
  et~al.}]{perkins_transientproteinproteininteractions_2010}
Perkins, J.~R., Diboun, I., Dessailly, B.~H., et~al. (2010).
\newblock Transient protein-protein interactions: Structural, functional, and
  network properties.
\newblock \emph{Structure} 18, 1233--1243.
\newblock \doi{10.1016/j.str.2010.08.007}
\bibAnnoteFile{perkins_transientproteinproteininteractions_2010}

\bibitem[{Raffel et~al.(2023)Raffel, Shazeer, Roberts
  et~al.}]{raffel_exploringlimitstransfer_2023}
[Dataset] Raffel, C., Shazeer, N., Roberts, A., et~al. (2023).
\newblock Exploring the {{Limits}} of {{Transfer Learning}} with a {{Unified
  Text-to-Text Transformer}}.
\newblock \doi{10.48550/arXiv.1910.10683}
\bibAnnoteFile{raffel_exploringlimitstransfer_2023}

\bibitem[{Rao et~al.(2020)Rao, Meier, Sercu
  et~al.}]{rao_transformerproteinlanguage_2020}
Rao, R., Meier, J., Sercu, T., et~al. (2020).
\newblock Transformer protein language models are unsupervised structure
  learners.
\newblock In \emph{International {{Conference}} on {{Learning
  Representations}}}
\bibAnnoteFile{rao_transformerproteinlanguage_2020}

\bibitem[{Rao et~al.(2014)Rao, Srinivas, Sujini
  et~al.}]{rao_proteinproteininteractiondetection_2014}
Rao, V.~S., Srinivas, K., Sujini, G.~N., et~al. (2014).
\newblock Protein-{{Protein Interaction Detection}}: {{Methods}} and
  {{Analysis}}.
\newblock \emph{Int J Proteomics} 2014, 147648.
\newblock \doi{10.1155/2014/147648}
\bibAnnoteFile{rao_proteinproteininteractiondetection_2014}

\bibitem[{Rossi et~al.(2023)Rossi, Charpentier, Di~Giovanni
  et~al.}]{rossi_edgedirectionalityimproves_2023}
[Dataset] Rossi, E., Charpentier, B., Di~Giovanni, F., et~al. (2023).
\newblock Edge {{Directionality Improves Learning}} on {{Heterophilic Graphs}}
\bibAnnoteFile{rossi_edgedirectionalityimproves_2023}

\bibitem[{Scannell et~al.(2012)Scannell, Blanckley, Boldon
  et~al.}]{scannell_diagnosingdeclinepharmaceutical_2012}
Scannell, J.~W., Blanckley, A., Boldon, H., et~al. (2012).
\newblock Diagnosing the decline in pharmaceutical {{R}}\&{{D}} efficiency.
\newblock \emph{Nat Rev Drug Discov} 11, 191--200.
\newblock \doi{10.1038/nrd3681}
\bibAnnoteFile{scannell_diagnosingdeclinepharmaceutical_2012}

\bibitem[{Scarselli et~al.(2009)Scarselli, Gori, Tsoi
  et~al.}]{scarselli_graphneuralnetwork_2009}
Scarselli, F., Gori, M., Tsoi, A.~C., et~al. (2009).
\newblock The {{Graph Neural Network Model}}.
\newblock \emph{IEEE Trans. Neural Networks} 20, 61--80.
\newblock \doi{10.1109/tnn.2008.2005605}
\bibAnnoteFile{scarselli_graphneuralnetwork_2009}

\bibitem[{Scott et~al.(2016)Scott, Bayly, Abell
  et~al.}]{scott_smallmoleculesbig_2016}
Scott, D.~E., Bayly, A.~R., Abell, C., et~al. (2016).
\newblock Small molecules, big targets: Drug discovery faces the
  protein--protein interaction challenge.
\newblock \emph{Nat Rev Drug Discov} 15, 533--550.
\newblock \doi{10.1038/nrd.2016.29}
\bibAnnoteFile{scott_smallmoleculesbig_2016}

\bibitem[{Sledzieski et~al.(2021)Sledzieski, Singh, Cowen
  et~al.}]{sledzieski_dscripttranslatesgenome_2021}
Sledzieski, S., Singh, R., Cowen, L., et~al. (2021).
\newblock D-{{SCRIPT}} translates genome to phenome with sequence-based,
  structure-aware, genome-scale predictions of protein-protein interactions.
\newblock \emph{cels} 12, 969--982.e6.
\newblock \doi{10.1016/j.cels.2021.08.010}
\bibAnnoteFile{sledzieski_dscripttranslatesgenome_2021}

\bibitem[{Trabuco et~al.(2012)Trabuco, Betts, and
  Russell}]{trabuco_negativeproteinprotein_2012}
Trabuco, L.~G., Betts, M.~J., and Russell, R.~B. (2012).
\newblock Negative protein--protein interaction datasets derived from
  large-scale two-hybrid experiments.
\newblock \emph{Methods} 58, 343--348.
\newblock \doi{10.1016/j.ymeth.2012.07.028}
\bibAnnoteFile{trabuco_negativeproteinprotein_2012}

\bibitem[{Vaswani et~al.(2017)Vaswani, Shazeer, Parmar
  et~al.}]{vaswani_attentionallyou_2017}
Vaswani, A., Shazeer, N., Parmar, N., et~al. (2017).
\newblock Attention {{Is All You Need}}.
\newblock \emph{arXiv:1706.03762 [cs.CL]}
\bibAnnoteFile{vaswani_attentionallyou_2017}

\bibitem[{Veli{\v c}kovi{\'c} et~al.(2018)Veli{\v c}kovi{\'c}, Cucurull,
  Casanova et~al.}]{velickovic_graphattentionnetworks_2018}
Veli{\v c}kovi{\'c}, P., Cucurull, G., Casanova, A., et~al. (2018).
\newblock Graph {{Attention Networks}}.
\newblock \emph{arXiv:1710.10903 [stat.ML]} \doi{10.48550/arXiv.1710.10903}
\bibAnnoteFile{velickovic_graphattentionnetworks_2018}

\bibitem[{Vidal et~al.(2011)Vidal, Cusick, and
  Barab{\'a}si}]{vidal_interactomenetworkshuman_2011}
Vidal, M., Cusick, M.~E., and Barab{\'a}si, A.-L. (2011).
\newblock Interactome {{Networks}} and {{Human Disease}}.
\newblock \emph{Cell} 144, 986--998.
\newblock \doi{10.1016/j.cell.2011.02.016}
\bibAnnoteFile{vidal_interactomenetworkshuman_2011}

\bibitem[{Wodak and Janin(1978)}]{wodak_computeranalysisproteinprotein_1978}
Wodak, S.~J. and Janin, J. (1978).
\newblock Computer analysis of protein-protein interaction.
\newblock \emph{J Mol Biol} 124, 323--342.
\newblock \doi{10.1016/0022-2836(78)90302-9}
\bibAnnoteFile{wodak_computeranalysisproteinprotein_1978}

\bibitem[{Xu et~al.(2019)Xu, Hu, Leskovec et~al.}]{xu_howpowerfulare_2019}
Xu, K., Hu, W., Leskovec, J., et~al. (2019).
\newblock How {{Powerful}} are {{Graph Neural Networks}}?
\newblock \emph{arXiv:1810.00826 [cs.LG]}
\bibAnnoteFile{xu_howpowerfulare_2019}

\bibitem[{Zhou et~al.(2022)Zhou, Wang, Jin
  et~al.}]{zhou_graphneuralnetwork_2022}
Zhou, H., Wang, W., Jin, J., et~al. (2022).
\newblock Graph {{Neural Network}} for {{Protein}}--{{Protein Interaction
  Prediction}}: {{A Comparative Study}}.
\newblock \emph{Molecules} 27, 6135.
\newblock \doi{10.3390/molecules27186135}
\bibAnnoteFile{zhou_graphneuralnetwork_2022}

\end{thebibliography}

\end{document}